%% file: main.tex
\documentclass{article}

\PassOptionsToPackage{numbers, compress}{natbib}

\usepackage[main, final]{neurips_2026}

\usepackage{microtype}
\usepackage{graphicx}
\graphicspath{{figures/}}
\usepackage{subcaption}
\usepackage{booktabs}
\usepackage{float}
\usepackage[dvipsnames]{xcolor}
\usepackage[pagebackref=true,colorlinks,citecolor=RoyalBlue,urlcolor=RoyalBlue,bookmarks=false]{hyperref}
\usepackage{colortbl}
\usepackage[normalem]{ulem}
\usepackage{enumitem}

\usepackage{amsmath}
\usepackage{amssymb}
\usepackage{mathtools}
\usepackage{amsthm}
\usepackage{multirow}
\usepackage{xspace}
\usepackage{stfloats}
\usepackage{tikz}
\usepackage{adjustbox}

\usepackage[utf8]{inputenc}
\usepackage[T1]{fontenc}
\usepackage{url}
\usepackage{amsfonts}
\usepackage{nicefrac}
\usepackage{wrapfig}

\newcommand{\eg}{\textit{e}.\textit{g}.~}

\newcommand{\ours}{VidVec\xspace}
\usepackage[capitalize,noabbrev]{cleveref}

\theoremstyle{plain}

\theoremstyle{definition}

\theoremstyle{remark}

\usepackage[textsize=tiny]{todonotes}

\title{VidVec: Unlocking Video MLLM Embeddings for Video-Text Retrieval}

\author{\vspace{-0.6em}
  Issar Tzachor \hspace{1.5cm}
  Dvir Samuel \hspace{1.5cm}
  Rami Ben-Ari \\
  \\
  \vspace{-0.85em}
  OriginAI
}
\begin{document}

\maketitle

\vspace{-2.5mm}
\begin{abstract}
\vspace{-2.5mm}
Recent studies have adapted generative Multimodal Large Language Models (MLLMs) into embedding extractors for text--vision tasks, typically by fine-tuning them to produce universal representations. However, their performance on video--text retrieval often remains below that of dedicated Video Foundation Models. In this paper, we revisit the video--text retrieval capabilities of MLLMs.
We first analyze the zero-shot retrieval capabilities of pretrained MLLMs and show that they already encode substantial retrieval-relevant information: combining intermediate-layer embeddings with a calibrated MLLM head yields strong performance without any training.
To further exploit these capabilities, we introduce a lightweight text-based optimization strategy that requires no paired multimodal data.
Our method uses dense video captions as rich textual surrogates for video semantics and short summaries as compact, query-like views, showing that a carefully designed text-alignment objective can improve multimodal alignment through the shared MLLM backbone. We backup these findings through in-depth analysis.
Remarkably, our lightweight text-only optimization outperforms its counterpart trained with paired video–text supervision, which is substantially more computationally expensive. Furthermore, our method outperforms existing approaches, often by a substantial margin, and achieves state-of-the-art performance across common video-text retrieval benchmarks.
More broadly, our findings highlight the untapped retrieval capabilities of pretrained MLLMs and establish text-only optimization as a more efficient and effective substitute for large-scale visual supervision.
\href{https://iyttor.github.io/VidVec/}{Project page}

\end{abstract}

\vspace{-2mm}
\section{Introduction}
\vspace{-2mm}

Multimodal Large Language Models (MLLMs) have recently emerged as a dominant paradigm in vision–language understanding, demonstrating strong performance across tasks such as captioning \cite{videoCaptioning}, visual question answering \cite{openflamingoVQA}, and even visual mathematical reasoning \cite{chen2025mathflow}. A common design represents visual input as sequences of visual tokens that are fed into a {\it generative} Large Language Model (LLM) which, after joint fine-tuning, enables open-world reasoning and instruction following over multimodal content. Extending this paradigm to videos has led to rapid progress in video-based MLLMs \cite{LLaVa-Video24, damonlpsg2024videollama2, VideoLLaMA3_2025, video-SALMONN2_25, Keye-VL1.5_25}. Compared to images, videos introduce richer content and temporal dynamics, substantially increasing the challenge of representation learning, reasoning, and retrieval.

As MLLMs become increasingly widespread, multimodal representation learning has attracted growing research interest.
Recent large Video Foundation Models (VFMs) have pushed zero-shot and transfer performance by scaling video-text pretraining.
InternVideo2 \cite{Internvideo2_ECCV2024} has pretrained on $\sim$100M video-text pairs to reach superior performance, while VideoPrism \cite{VideoPrism} contrastively trained with $\sim$600M pairs.
Recent studies suggest that simply scaling video–text pretraining does not uniformly improve performance \cite{dataScalingIsssue1,dataScalingIssue2,DataScaleLimit_ICML25}, motivating alternative approaches that focus on representation quality and task-related alignment \cite{green2026find, effovpr_ICLR2025}, rather than relying solely on ever-larger corpora.

Recently, an emerging line of work has investigated whether MLLMs can serve as representation learners across vision--text modalities, reporting impressive results on image retrieval~\cite{GME_MLLM_Retrieval,B3_2025,kong2025UNITE,LAMRA_CVPR25,UniME-V2_AAAI26,xiao2026metaembed,cui2026think}. Yet, video is often not the primary focus, and reported video-retrieval performance often remains below that of dedicated Video Foundation Models (VFMs)~\cite{Internvideo2_ECCV2024,UMT_CVPR2022,LanguageBind_ICLR2024,LAMRA_CVPR25,lan2025llave}.

\begin{figure}[t]
\vspace{-10mm}
  \centering
\includegraphics[width=0.95\linewidth]{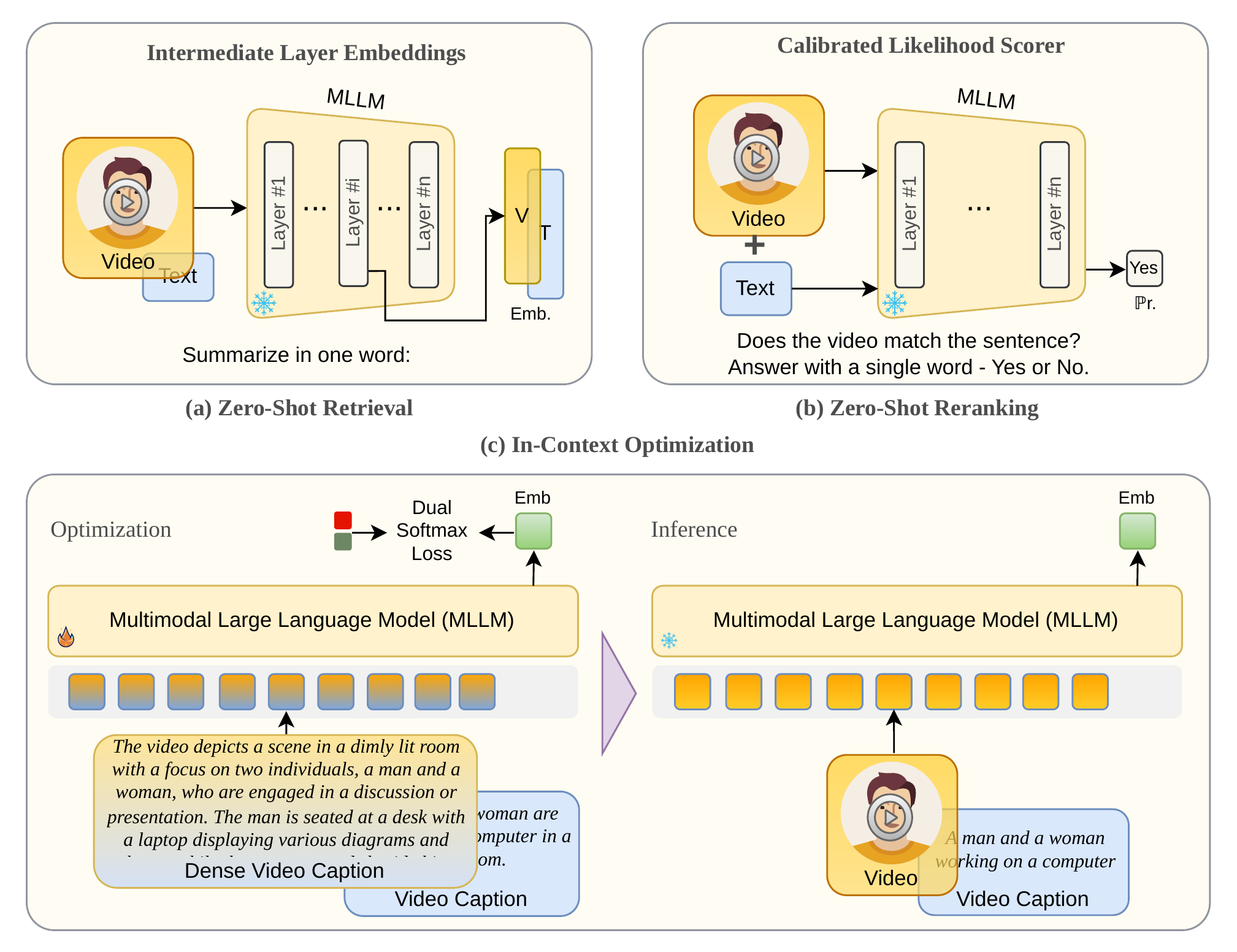}

  \captionsetup{font=small}
  \caption{An overview of VidVec. (a) \textbf{Zero-shot retrieval:} extract video and text embeddings from an intermediate MLLM layer for initial ranking. (b) \textbf{Zero-shot reranking:} leverage the calibrated MLLM head for pairwise scoring to rerank top-$K$ candidates. (c) \textbf{In-context optimization:} {\it lightweight} model alignment using only $\sim$60K text-only pairs for embedding extraction via a text-to-text mapping from dense video captions to short summaries, designed to mirror the video-text inference setup - dense caption serves as a surrogate for the full video content ({\it yellow}) and a short summary acts as a compact anchor closer to retrieval queries ({\it blue}).
}

  \label{fig:main}
  \vspace{-4mm}
\end{figure}

In this work, we revisit pretrained MLLMs as embedding extractors \emph{specifically} for video--text retrieval.
Although video MLLMs are trained primarily for generative video-language understanding, we show that their hidden representations already contain substantial retrieval-relevant structure.
As illustrated in Fig.~\ref{fig:main}, we first expose this structure in a purely zero-shot setting by extracting video and text embeddings from an intermediate MLLM layer. This yields strong initial retrieval performance without any training (Fig.~\ref{fig:main}a). We then reuse the MLLM generative head as a calibrated pairwise relevance scorer: given the top retrieved candidates, the model is prompted to answer whether a video matches a sentence, and the likelihood of an affirmative answer is used for reranking (Fig.~\ref{fig:main}b).
This simple two-stage pipeline already outperforms several trained MLLM embedders on standard video--text retrieval benchmarks, suggesting that retrieval-relevant video--text alignment is already encoded in pretrained video MLLMs.

To further unlock these encoded capabilities, we introduce a lightweight \emph{text-only} in-context optimization scheme that refines the embedding space without visual supervision.
A few prior MLLM embedders incorporated generic text supervision such as NLI~\cite{bowman-etal-2015-large} to induce retrieval-friendly representations~\cite{E5-V_2024,LAMRA_CVPR25}. We instead propose an in-context text-only optimization paradigm, where the supervision is constructed from textual surrogates of the target multimodal context. For video--text retrieval, we instantiate this paradigm with a video-grounded text-to-text proxy that aligns dense video captions with short summaries, tailoring the optimization signal to the target retrieval setting.
Dense video captions serve as rich textual surrogates for video semantics, while short summaries act as compact, query-like views of the same content.
Optimizing the MLLM to align these two textual views shapes the shared embedding space to better support video--text retrieval, yielding effective video--text representations at inference time.
Notably, using only $\sim$60K text-only pairs, our method not only surpasses direct video--text fine-tuning on the underlying videos but also outperforms MLLM embedders and VFMs trained on orders of magnitude more video--text data.

The key contributions of this work are:
\vspace{-1mm}
\begin{enumerate}
    \item
    We introduce an effective zero-shot scheme that
    turns off-the-shelf Video MLLM into a competitive video-text retriever.
    \vspace{-0.7mm}
    \item
    We present an efficient in-context optimization strategy that maps dense video captions to short summaries, enabling task-related embedding learning {\it without} visual supervision.
    \vspace{-0.7mm}

    \item We show that our text-only optimization is not only substantially more efficient than paired video--text supervision, but also yields stronger retrieval performance.
    \item
    We provide an empirical analysis of our in-context text-only optimization, showing that it induces a reorganization of the embedding space that improves video--text alignment.
    \vspace{-0.7mm}
    \item
    Without any fine-tuning beyond text, our method achieves state-of-the-art performance on common video retrieval benchmarks.
\end{enumerate}

\section{Related Work}
\vspace{-1mm}

{\bf Vision--language embeddings for retrieval.}
Early vision--language models (VLMs) such as CLIP~\cite{CLIP_ICML2021}, ALIGN~\cite{ALIGN_ICML2021}, BLIP~\cite{BLIP_PMLR_2022}, and SigLIP~\cite{SigLIP_ICCV2023} established a powerful dual-encoder paradigm for cross-modal retrieval by aligning image and text representations with contrastive objectives. This recipe has also been extended to video embedding and retrieval~\cite{Clip4clip2021,X-CLIP_ACMM2022,VideoCLIP2021}. However, dual encoders are less natural for interleaved multimodal inputs, instruction-conditioned retrieval, and video semantics that depend on temporal dynamics, motivating approaches based on instruction-following generative backbones and deeper multimodal fusion.

{\bf Video–text retrieval and video representation learning.}
Video retrieval has traditionally been addressed by video–language models that explicitly encode spatiotemporal structure and are trained on large-scale video–text corpora.
Recent Video Foundation Models (VFMs) substantially scale pretraining data and objectives often using huge datasets for training, for the aligned embedding task. For example, InternVideo \cite{InternVideo2022} reports pretraining on roughly 12M video clips spanning multiple domains, HowTo100M trains on nearly 136M short clips \cite{Howto100M_ICCV2019}. InternVideo2 \cite{Internvideo2_ECCV2024} further scales the data regime to 50M video–text pairs and another 50M
multimodal (video–audio–speech–text) pairs. Other methods, \eg, UMT \cite{UMT_CVPR2022}, LanguageBind \cite{LanguageBind_ICLR2024} and VideoPrism \cite{VideoPrism} present state-of-the-art results while building on hundreds of millions of paired samples (video-text), often in addition to millions of image-text pairs.

These approaches typically rely on video-specialized encoders, explicit temporal modeling, and large-scale video--text supervision. In contrast, our work studies {\em multimodal LLMs as embedders} for video understanding, demonstrated through video retrieval, focusing on (i) whether pretrained video MLLMs already encode retrieval-relevant video--text structure, and (ii) how far video retrieval can be improved using only {\em text-only} supervision.

{\bf Multimodal LLMs as embedders and token-based embedding learning.}
UniIR \cite{uniIR_2024} introduced instruction-guided multimodal retrieval with image-text paired post-training while MagicLens \cite{Magiclens_2024} explored self-supervised instruction-conditioned {\it image} retrieval using web-mined image pairs and synthesized instructions. Both methods are based on VLM backbones \eg BLIP \cite{BLIP_PMLR_2022} or CLIP.

In parallel, VLM2Vec \cite{VLM2Vec_ICLR25} proposed an instruction-guided contrastive framework that converts MLLMs into embedding models by LoRA finetuning on large scale MMEB dataset \cite{VLM2Vec_ICLR25}. VLM2Vec-V2 \cite{VLM2Vec-V2_2025} extends this direction by explicitly incorporating videos and visual documents into training.

A growing line of work repurposes multimodal LLMs (MLLMs/LMMs) as embedding models by extracting fixed-dimensional hidden-state representations with carefully designed prompts.
E5-V \cite{E5-V_2024} showed that prompt-based “semantic compression” (e.g., asking the model to produce a one-word summary) can create aligned image-text (multimodal) embeddings,  even without fine-tuning; critically, E5-V further demonstrated that {\em single-modality training only on text pairs} from Natural Language Inference (NLI) dataset~\cite{bowman-etal-2015-large} can generate reasonably aligned multimodal embeddings with low training cost, avoiding the need for image–text training data.

LamRA \cite{LAMRA_CVPR25} follows this paradigm, building on large multimodal models and showed significantly improved results with additional image-text paired training. Technically, it uses an “Explicit One-word Limitation” prompt of the form “Summarize … in one word: {\tt <emb>}” and takes the last hidden state immediately preceding the {\tt <emb>} token as the embedding. It further combines language-only pretraining with multimodal instruction tuning, and uses a separately trained reranking module.

CARE \cite{xu2024carebench} pretrains an MLLM embedder by generating detailed video captions, followed by training on general text pairs. In contrast, UNITE \cite{kong2025UNITE} trains on modality-specific pairs (text–text, text–image, and text–video), suggesting that generation-based pretraining does not always improve performance.

Prior MLLM embedders use prompt-based token readout and text-only supervision from generic textual pairs (NLI), but largely overlook the strong zero-shot retrieval signal already present in pretrained MLLMs. In contrast, we first expose this capability through intermediate-layer embeddings, and then introduce text-only \emph{in-context} supervision as a task-grounded alternative, where training pairs are derived from textual views of the target context itself. We instantiate this paradigm for video--text retrieval by aligning dense captions, which proxy video content, with summaries, which provide compact query-like views. This video-specific text-only signal reduces the query--video semantic mismatch, offering an efficient alternative to large-scale video--text contrastive training.

Beyond token prompting and readout design, recent work has improved MLLM-based embeddings through stronger training recipes and richer interaction mechanisms. B3~\cite{B3_2025} studies efficient contrastive training with improved batch and negative construction, while UniME-V2~\cite{UniME-V2_AAAI26} uses an ``MLLM-as-a-judge'' to assign soft matching scores for hard negative mining. UME-R1~\cite{UME-R1_2025} explores reasoning-driven {\em generative} embeddings with SFT and RL. Qwen3-VL-Embedding~\cite{qwen3vlembedding} combines multimodal embedding learning with a dedicated reranker, Think-Then-Embed~\cite{cui2026think} conditions embeddings on generated reasoning traces and MetaEmbed~\cite{xiao2026metaembed} introduces compact multi-vector embeddings with flexible late interaction.
Our work is complementary: we focus on video--text retrieval and introduce a lightweight dense-caption--to--summary objective that improves video-aligned embeddings without visual supervision.

\vspace{-3mm}
\section{Method}
\vspace{-3mm}

\label{sec:method}
We introduce \ours, which adapts pretrained generative MLLMs to video--text retrieval while avoiding visual supervision. As illustrated in Fig.~\ref{fig:main}, \ours consists of three components. First, in the zero-shot embedding stage (Fig.~\ref{fig:main}a), we extract video and text representations from an intermediate layer of an off-the-shelf video MLLM. This exposes retrieval-relevant structure already present in the pretrained model, and provides an initial ranking without any training.

Second, in zero-shot reranking (Fig.~\ref{fig:main}b), we reuse the MLLM language-model head as a calibrated relevance scorer: for each top-ranked candidate pair, the model is prompted to answer whether the video matches the sentence, and the likelihood of an affirmative answer is used for reranking.

Finally, building on this zero-shot capability, we introduce a lightweight text-only \emph{in-context} optimization objective (Fig.~\ref{fig:main}c). Instead of relying on paired video--text supervision, we use two textual views from the same video context: a dense caption that describes the video in detail, and a short summary that captures its salient semantics in a query-like form. We then align matched dense-caption--summary pairs against mismatched pairs, using the dense caption as a textual proxy for the video and the summary as a proxy for the retrieval query. This in-context text-only optimization improves video--text alignment without direct visual supervision.

We next formalize the retrieval setup, describe the zero-shot embedding and reranking components, and then motivate our \emph{in-context} optimization approach.

We follow the standard video--text retrieval setting, using a generative Multimodal Large Language Model (MLLM). Given a query $q$ (either text or video) and a candidate pool $\Omega={c_1,\dots,c_N}$ of size $N$, where $c_i$ belongs to the other modality of $q$, we encode the query and each candidate with an embedding extractor
\(f_{\theta}(\cdot)\in\mathbb{R}^d\), where \(\theta\) denotes the parameters of the MLLM. We score each candidate by cosine similarity and rank $\Omega$ accordingly.
For second stage re-ranking, we retain the top-$K$ candidates, denoted by \(\mathcal{R}_K\), and apply a reranking function over this reduced set to obtain the final ranked list, \(
\mathcal{R}_{\mathrm{final}}=\Phi_{\mathrm{rerank}}(q,\mathcal{R}_K),
\)
where \(\mathcal{R}_{\mathrm{final}}\) is the final ordered set of candidates produced by the reranking stage.

\subsection{Embedding Extraction}
\label{sec:feature_extraction}
Multimodal large language models (MLLMs) typically comprise a vision encoder, a projector that maps visual features into the language-model input space, and a language model. After projection, visual tokens and text tokens are processed by the shared multimodal backbone \cite{LLaVa-Video24, Qwen2.5VL2205, wang2024qwen2,VideoLLaMA3_2025}

To extract embeddings from MLLMs, we follow prior work \cite{E5-V_2024, LAMRA_CVPR25} and adopt an \emph{Explicit One-word Limitation} (EOL) prompting scheme, in which the prompt ends with the instruction to respond \texttt{``in one word''} followed by a dedicated \texttt{<emb>} token. We take as the embedding the hidden state immediately preceding \texttt{<emb>} and refer to it as \texttt{<emb-1>}.

\subsection{MLLM Head as a Calibrated
Likelihood Scorer}
\label{sec:MLLM-head-scorer}
For the second stage of re-ranking, we reuse the language model head as a calibrated likelihood scorer. Given the reduced candidate set $\mathcal{R}_K$, each query--candidate pair $(q,c_i)$ is evaluated independently by prompting the model with a binary relevance question. Concretely, we extend the EOL prompting scheme with the instruction \texttt{``Respond in a single word - Yes or No.''}. We then define the re-ranking score for each candidate $c_i$ as the likelihood of the affirmative tokens \texttt{Yes}/\texttt{yes}, $S_{\text{rank}}(q,c_i)=P_\theta(\texttt{Yes}\mid q,c_i)$, where $S_{\text{rank}}$ is the relevance similarity score used for final ranking. This procedure requires $K$ forward passes and induces the re-ranked list, where candidates in $\mathcal{R}_K$ are ordered by decreasing $S_{\text{rank}}$. Fig.~\ref{fig:main}b describes this setting.

\subsection{In-context Text-only Optimization}
\label{sec:in-context-optimization}
In this section, we motivate our text-only in-context optimization as a proxy objective for text--video retrieval while empirical support is provided later in Sec.~\ref{sec:analysis}.
Fig.~\ref{fig:main}c illustrates the setup.

Let us consider a triplet $\{v,t_d,t_s\}$ denoting a video sequence, its corresponding dense caption and short summary respectively.
Our ultimate retrieval objective is to align a textual query \(t_q\) with its corresponding video embedding, i.e.,
\[
\min_{\theta} D\!\big(f_\theta(t_q), f_\theta(v)\big),
\]
where \(D(\cdot,\cdot)\) is a distance measure in the embedding space.
Since our optimization is text-only, we instead employ the following surrogate objective, aligning dense-captions with short-summaries:
\[
\min_{\theta} D\!\big(f_\theta(t_s), f_\theta(t_d)\big).
\]

\vspace{-1mm}
Our motivation is based on two observations. First, in our zero-shot analysis (see Sec.~\ref{sec:analysis} and Appendix Sec.~\ref{sec:further-in-context}), embeddings of dense captions are better aligned with the paired video embeddings than embeddings of short summaries. We therefore treat the dense caption \(t_d\) as a stronger textual proxy for the video's semantic content in the pretrained embedding space.
Second, a short summary compactly captures the video's salient semantics.
We therefore treat $t_s$ as analogous to a query $t_q$.

Since the pretrained backbone already exhibits strong zero-shot video--text retrieval capabilities, as demonstrated in Sec.~\ref{sec:zs-section}, the objective should not be interpreted as learning a literal mapping \(t_s \longleftrightarrow t_d\). Rather, it serves as a proxy for exposing and refining a shared latent representation that is already encoded in the model.

Our text-only optimization is designed to improve video-text alignment by making the textual representations induced by the same video semantics more consistent. The effect on video representations is indirect but well motivated. Let \(z_v\) denote the underlying semantic content of video \(v\). We treat its dense caption \(t_d\) and short summary \(t_s\) as textual realizations of this content: \(t_d\) preserves richer visual and temporal detail, while \(t_s\) provides a compact abstraction.
Thus, both correspond to the same latent video semantics \(z_v\).
We optimize $\theta$ such that, for matched pairs $(t_s^i,t_d^i)$, the similarity is maximized relative to mismatched pairs $(t_s^i,t_d^j)$ with $i\neq j$. This encourages invariance to caption length, phrasing, and descriptive granularity, while preserving the shared semantics induced by the same video semantics.

Crucially, both text and video are encoded by the same shared backbone $f_\theta$, while the vision encoder and its projector are frozen.
Thus, updating $\theta$ to better align the two textual views associated with the same latent video semantics jointly reshapes the induced representation of the corresponding video.

We adopt Dual-Softmax Loss \cite{cheng2021improving_DSL} as the objective for our in-context optimization. This loss is commonly used in video--text contrastive training \cite{Webvid_ICCV2023,Clip4clip2021, InternVideo2022}. Given a minibatch of paired dense captions and short summaries, we compute the cross-view similarity matrix between \(f_\theta(t_s^i)\) and \(f_\theta(t_d^j)\). We then apply softmax normalization along both axes of this matrix, yielding conditional probability distributions for dense-caption-to-summary and summary-to-dense-caption retrieval. These two distributions are multiplicatively combined to emphasize pairs with high mutual matching score under both retrieval directions, and the objective is applied symmetrically.

Overall, this yields a lightweight adaptation scheme that operates entirely in text space using in-context captions derived from the same video. Compared with full video--text fine-tuning, it is far more efficient, avoids the need for paired visual supervision, and provides an effective way to improve modality alignment without directly retraining on video inputs.

\vspace{2pt}

\section{Evaluation}
In this section, we compare our approach against several state-of-the-art MLLM embedders and VFMs. For MLLMs, we evaluate seven methods, including UNITE \cite{kong2025UNITE}, MMRet-v1.5 (BGE-VL) \cite{zhou2024megapairs}, VLM2Vec-V2 \cite{VLM2Vec-V2_2025}, and the recent UniME-V2 \cite{UniME-V2_AAAI26}. For VFMs, we compare to seven methods, including InternVideo2-6B \cite{Internvideo2_ECCV2024}, VideoPrism-g \cite{VideoPrism}, and PE-Core-G \cite{bolya2025perception}.

We conduct an evaluation with four common video-text retrieval benchmarks - MSR-VTT \cite{xu2016msrvtt}, MSVD \cite{chen2011collecting_msvd}, VATEX \cite{wang2019vatex}, and DiDeMo \cite{anne2017localizing_didemo}.
Additional information on the benchmarks is provided in the Appendix.
Following prior {\it zero-shot} protocol~\cite{InternVideo2022,Internvideo2_ECCV2024,VideoPrism}, we ensure models are not exposed to benchmark training sets and compare only MLLM embedders with confirmed non-exposure~\cite{seed1-6, qwen3vlembedding}.

VLM2Vec-V2 \cite{VLM2Vec-V2_2025} introduced the MMEB-V2 benchmark, which spans multiple video tasks. Here we focus on video–text retrieval and use the standard datasets and splits adopted by recent VFMs \cite{VideoPrism,Internvideo2_ECCV2024, bolya2025perception} to enable direct comparison with current state-of-the-art methods. We still compare against VLM2Vec-V2, but report results on the established retrieval splits. While MMEB-V2 includes some existing retrieval datasets, it does not evaluate the V2T setting and modifies several test splits, making direct comparisons to current state-of-the-art methods difficult. For completeness, in the Appendix we compare against the MMEB-V2 reported results on the overlapping video-retrieval datasets, where \ours achieves superior performance.

We present three variants of our method: (1) a zero-shot two-stage baseline (\ours-ZS), which involves {\it no training}, (2) a one-stage in-context optimization approach (\ours-O), and (3) a two-stage (\ours) with optional reranking.

\subsection{Implementation Details}
For our in-context optimization,
we leverage the publicly available VideoUFO dataset \cite{VideoUFO_2025}, which contains over 1.09M video clips, each paired with both brief and dense video captions. From this corpus, we sample 60K data entries and fine-tune the model to map dense descriptions to short summaries.
We fine-tune a lightweight LoRA  \cite{mangrulkar2022peft}, and use a batch size of 288 pairs distributed across 4 B200 GPUs, which completes in under $30$ minutes. For reranking, we set \(K=5\) - processing each query--candidate pair incurs a latency of 121 ms on a single B200 GPU. We adopt VideoLLaMA3 \cite{VideoLLaMA3_2025} as our backbone for its superior zero-shot performance (see Appendix). For the zero-shot variant \ours-ZS, we use embeddings extracted from layer $24$, whereas for the optimized variant \ours-O we use the final-layer embeddings.
Videos are sampled at 2 FPS (matching the Qwen2-VL default) and capped at 180 frames. Following prior VFM work, we also apply dual-softmax at inference to boost performance \cite{cheng2021improving_DSL, InternVideo2022,Internvideo2_ECCV2024,bolya2025perception}.
To ensure a fair comparison, we re-run competing MLLM embedders, where applicable, using the same FPS and a dual-softmax calibrated scoring. This often improves performance relative to the numbers reported in the original papers. For VFMs, we report the published results. Additional implementation details are provided in the Appendix.

\subsection{Zero-shot Performance}
\label{sec:zs-section}

\begin{wraptable}{r}{0.41\linewidth}
\vspace{-4.2mm}
\centering
\captionsetup{type=table,font=small}
\caption{Text-to-Video (T2V) Retrieval - our two-stage \ours-ZS ({\it zero-shot}) vs. state-of-the-art MLLM embedders (7B size).}
\label{tab:zs-t2v-results}
\small
\setlength{\tabcolsep}{6pt}
\resizebox{0.41\textwidth}{!}{\begin{tabular}{l|c|c|c}
\toprule
Model & MSR-VTT & VATEX & DiDeMo \\
\midrule
LamRA        & 48.9 & 61.4 & 46.3 \\
VLM2Vec      & 41.6 & 53.6 & 37.7 \\
MMRet-v1.5   & 45.2 & 57.7 & 45.1 \\
B3           & 46.7 & 57.3 & 38.7 \\
VLM2Vec-V2   & 41.0 & 52.9 & 41.7 \\
UNITE        & 46.5 & 45.7 & 43.5 \\
UniME-V2     & 40.6 & 52.0 & 36.5 \\
\midrule
{\color{gray} VidVec-ZS-Emb}    & {\color{gray} 42.1} & {\color{gray} 54.7} & {\color{gray} 38.8} \\
VidVec-ZS                & \textbf{51.2}       & \textbf{67.6}       & \textbf{53.1} \\
\bottomrule
\end{tabular}
}
\end{wraptable}

In Tab.~\ref{tab:zs-t2v-results}, we evaluate our zero-shot two-stage method, against state-of-the-art MLLM embedders {\it trained} to produce vision–language embeddings.
We report results on text-to-video (T2V) retrieval for MSR-VTT, VATEX and DiDeMo. While prior methods rely on contrastive training over vision--text pairs, \ours-ZS leverages an off-the-shelf MLLM by extracting embeddings from a fixed intermediate layer, followed by a re-ranking stage.
Despite requiring no training, \ours-ZS surpasses prior methods across all three benchmarks, achieving Recall@1 gains of \(+2.3\%\), \(+6.2\%\), and \(+6.8\%\) on MSR-VTT, VATEX, and DiDeMo, respectively. Notably, using intermediate-layer embeddings alone (denoted as \ours-ZS-Emb), already yields competitive performance and outperforms multiple prior methods.
This indicates that off-the-shelf generative MLLMs already encode well-aligned video--text representations internally, a property demonstrated here and not previously established in prior work \cite{E5-V_2024,LAMRA_CVPR25,UniME-V2_AAAI26}.

\begin{table}[t]
\centering
\caption{Text-to-Video (T2V) retrieval - VidVec-O ({\it Optimized Embedding}) vs.\ state-of-the-art MLLM embedders MLLM embedders (7B size) on four benchmarks. \ours-O uses only 60K \emph{text-only} in-context pairs, whereas prior methods are trained on $\times$10-$\times$100 larger scale of \emph{vision-text} data.}

\resizebox{\textwidth}{!}{\begin{tabular}{l|ccc|ccc|ccc|ccc}
\toprule
 & \multicolumn{3}{c|}{MSR-VTT} & \multicolumn{3}{c|}{MSVD} & \multicolumn{3}{c|}{VATEX} & \multicolumn{3}{c}{DiDeMo} \\
Model & R@1 & R@5 & R@10 & R@1 & R@5 & R@10 & R@1 & R@5 & R@10 & R@1 & R@5 & R@10 \\
\midrule
LamRA & 48.9 & 71.6 & 79.5 & 55.7 & 81.7 & 88.0 & 61.4 & 90.6 & 95.2 & 46.3 & 71.8 & 80.5 \\
VLM2Vec & 41.6 & 64.5 & 73.8 & 52.1 & 78.4 & 85.8 & 53.6 & 84.4 & 91.0 & 37.7 & 60.9 & 69.9 \\
MMRet-v1.5 & 45.2 & 67.5 & 75.9 & 49.5 & 76.6 & 83.5 & 57.7 & 87.5 & 93.5 & 45.1 & 70.7 & 78.8 \\
B3 & 46.7 & 67.6 & 75.5 & 53.8 & 80.0 & 86.7 & 57.3 & 86.5 & 92.3 & 38.7 & 62.6 & 70.7 \\
VLM2Vec-V2 & 41.0 & 63.6 & 71.9 & 48.2 & 75.4 & 83.1 & 52.9 & 83.2 & 90.4 & 41.7 & 66.5 & 75.1 \\
UNITE & 46.5 & 69.4 & 78.0 & 50.4 & 78.2 & 86.4 & 45.7 & 74.0 & 81.0 & 43.5 & 69.6 & 77.3 \\
UniME-V2 & 40.6 & 62.5 & 70.7 & 52.1 & 77.5 & 84.4 & 52.0 & 80.8 & 86.9 & 36.5 & 59.2 & 66.7 \\
\midrule
\textbf{VidVec-O} & \textbf{52.5} & \textbf{76.3} & \textbf{83.8} & \textbf{60.8} & \textbf{84.9} & \textbf{90.1} & \textbf{68.2} & \textbf{93.6} & \textbf{96.8} & \textbf{53.7} & \textbf{79.4} & \textbf{85.0} \\
\bottomrule
\end{tabular}}
\label{tab:united_t2v_retrieval}

\centering
\small
\caption{Video-to-Text (V2T) retrieval - VidVec-O ({\bf O}ptimized Embedding) vs.\ state-of-the-art MLLM embedders (7B size) on four benchmarks.
}
\resizebox{\textwidth}{!}{\begin{tabular}{l|ccc|ccc|ccc|ccc}
\toprule
 & \multicolumn{3}{c|}{MSR-VTT} & \multicolumn{3}{c|}{MSVD} & \multicolumn{3}{c|}{VATEX} & \multicolumn{3}{c}{DiDeMo} \\
Model & R@1 & R@5 & R@10 & R@1 & R@5 & R@10 & R@1 & R@5 & R@10 & R@1 & R@5 & R@10 \\
\midrule
LamRA & 50.9 & 72.8 & 80.0 & 81.2 & 92.5 & 95.5 & 82.5 & 95.4 & 97.7 & 47.1 & 72.8 & 81.5 \\
VLM2Vec & 44.9 & 68.6 & 76.1 & 81.3 & 93.1 & 95.7 & 80.2 & 95.5 & 98.7 & 41.3 & 65.4 & 72.5 \\
MMRet-v1.5 & 47.9 & 70.9 & 77.7 & 77.2 & 90.9 & 93.4 & 81.1 & 95.7 & 97.7 & 46.1 & 73.1 & 80.7 \\
B3 & 48.7 & 70.5 & 77.9 & 83.1 & 95.2 & 97.2 & 83.1 & 96.4 & 98.5 & 43.3 & 65.3 & 72.5 \\
VLM2Vec-V2 & 42.9 & 64.8 & 72.2 & 82.1 & 95.1 & 97.5 & 79.9 & 95.9 & 98.2 & 44.2 & 68.0 & 76.1 \\
UNITE & 45.2 & 70.3 & 79.3 & 76.1 & 91.3 & 94.6 & 78.9 & 95.3 & 98.1 &  40.3 & 68.7 & 78.1 \\
UniME-V2 & 46.3 & 65.7 & 72.4 & 81.8 & 93.4 & 95.5 & 81.5 & 96.1 & 98.4 & 38.7 & 61.0 & 69.0 \\
\midrule
\textbf{VidVec-O} & \textbf{54.9} & \textbf{77.5} & \textbf{84.1} & \textbf{85.7} & \textbf{95.5} & \textbf{97.8} & \textbf{89.6} & \textbf{98.5} & \textbf{99.3} & \textbf{56.5} & \textbf{79.7} & \textbf{86.0} \\
\bottomrule
\end{tabular}}
\label{tab:united_v2t_retrieval}
\vspace{-5mm}
\end{table}

\subsection{Comparison with State-of-the-Art}
In this section, we evaluate our {\it lightweight} in-context optimization approach for extracting output embeddings. We assess the resulting embeddings on four standard video–text retrieval datasets, comparing with state-of-the-art MLLM embedders and VFMs under the same evaluation protocol.

We compare the embeddings extracted by our optimized model, \ours-O, against recent MLLM embedders. Importantly, \ours-O uses no reranking. In Tab.~\ref{tab:united_t2v_retrieval} and  Tab.~\ref{tab:united_v2t_retrieval}, we report Text-to-Video (T2V) and Video-to-Text (V2T) results respectively. Across both retrieval directions, \ours-O consistently surpasses existing MLLM embedders, often by a substantial margin. The results highlight the effectiveness of our approach, yielding notable Recall@1 improvements, including V2T +6.5\% on VATEX and +9.4\% on DiDeMo. This demonstrates that our in-context optimization can effectively elicit the model’s internal knowledge using relatively small amounts of textual data, yielding better-aligned video-text embeddings than prior methods trained on millions of vision–text pairs.

While text-based optimization leads to performance gains, the calibrated head reranker in VidVec-ZS provides even larger improvements.
Moreover, applying this reranker to only a handful of VidVec-O top candidates further improves text-to-video retrieval, achieving state-of-the-art performance.

Tab.~\ref{tab:merged_t2v_v2t_r1} reports results obtained by adding a second-stage reranking on top of \ours-O. We compare Recall@1 performance against state-of-the-art Video Foundation Models (VFMs).
\begin{table}[t]
\centering
\small
\caption{Video-Text retrieval performance (Recall@1): \ours
vs. state-of-the-art Video Foundation Models (VFMs). Dashed entries indicate not reported results in the corresponding paper.}
\setlength{\tabcolsep}{5pt}
\resizebox{0.85\textwidth}{!}{\begin{tabular}{l c|cc|cc|cc|cc}
\toprule
\multirow{2}{*}{Method} & \multirow{2}{*}{\shortstack{V-T\\Pairs}}
& \multicolumn{2}{c|}{MSR-VTT} & \multicolumn{2}{c|}{MSVD} & \multicolumn{2}{c|}{VATEX} & \multicolumn{2}{c}{DiDeMo} \\
\cmidrule(lr){3-4}\cmidrule(lr){5-6}\cmidrule(lr){7-8}\cmidrule(lr){9-10}
&
& T2V & V2T & T2V & V2T & T2V & V2T & T2V & V2T \\
\midrule
Clip4Clip\cite{Clip4clip2021}       & n/a     & 32.0 & -    & 38.5 & -    & -    & -    & -    & - \\
ViCLIP\cite{VideoCLIP2021}          & n/a     & 42.4 & 41.3 & 49.1 & 75.1 & -    & -    & 18.4 & 27.9 \\
InternVideo-L\cite{InternVideo2022}   & 12M    & 40.7 & 39.6 & 43.4 & 67.6 & 49.5 & 69.5 & 31.5 & 33.5 \\
UMT-L\cite{UMT_CVPR2022}           & 10M    & 42.6 & 38.6 & 49.9 & 75.4 & -    & -    & 48.6 & 49.9 \\
LanguageBind\cite{LanguageBind_ICLR2024}    & 10M    & 44.8 & 40.9 & 53.9 & 72.0 & -    & -    & 39.9 & - \\
VideoPrism-g\cite{VideoPrism}    & 600M   & 52.7 & 51.7 & -    & -    & 62.5 & 77.1 & -    & - \\
PE-Core-G\cite{bolya2025perception}       & 22M    & 51.2 & 49.9 & 59.7 & 85.4 & -    & -    & -    & - \\
InternVideo2-6B\cite{Internvideo2_ECCV2024} & 100M   & 55.9 & 53.7 & 59.3 & 83.1 & \textbf{71.5} & 85.3 & 57.9 & \textbf{57.1} \\
\midrule
\ours & Text-Only & \textbf{56.4} & \textbf{54.9} & \textbf{61.6} & \textbf{85.7} & 70.5 & \textbf{89.6} & \textbf{62.3} & 56.5 \\
\bottomrule
\end{tabular}
}
\label{tab:merged_t2v_v2t_r1}
\end{table}

Regarding training data, Perception Encoder (PE-Core) \cite{bolya2025perception} and VideoPrism \cite{VideoPrism} used approximately 5.4B image-text (and 22M video–text pairs) and 1B image–text and 600M video–text pairs, including 36.1M manually labeled samples for training. InternVideo2-6B \cite{Internvideo2_ECCV2024}, which previously achieved state-of-the-art performance on most benchmarks, is trained on about 300M image–text pairs and 100M video–text pairs, with additional gains achieved from a second-stage re-ranking. \ours delivers state-of-the-art performance on the majority of evaluated benchmarks in both T2V and V2T retrieval. +1.2\% improvement on MSR-VTT V2T, and +1.2\% on MSVD T2V. On VATEX, \ours underperforms by -1.5\% on T2V but improves by +4.2\% on V2T, and on DiDeMo, while underperforming by -0.6\% on V2T, it improves by +3.9\% on T2V.

Although \ours uses a video-pretrained MLLM backbone, it is not explicitly trained for modality alignment or embedding extraction. This highlights the effectiveness of our lightweight in-context optimization and shows that strong video–text representations can be efficiently unlocked from a powerful MLLM. We further demonstrate generalizability across MLLM backbones in Appendix~\ref{app:backbone_generalization}.

\vspace{-3mm}
\section{Direct Video--Text Supervision and Ablations}
\label{sec:video-text-supervision}
\vspace{-3mm}

In Tab.~\ref{tab:video-text-supervision}, we compare our text-only optimization with direct video--text fine-tuning using the corresponding VideoUFO clips under the same training configuration. Our method substantially outperforms video--text training under matched compute and, remarkably, achieves superior performance even when the latter uses all training examples at approximately $20\times$ the training time. This suggests that textual supervision is not only more efficient but can also be more effective than direct visual supervision. Further details and discussion are provided in Appendix~\ref{app:video-text-supervision}.

\vspace{-1.5em}
\begin{table}[H]
\centering
\small
\captionsetup{font=small}
\caption{Text-only optimization vs. direct video--text fine-tuning.}
\label{tab:video-text-supervision}
\resizebox{\linewidth}{!}{
\begin{tabular}{llcccccccc}
\toprule
Supervision &
Training &
\# Train &
Training &
\multicolumn{3}{c}{MSR-VTT} &
\multicolumn{3}{c}{DiDeMo} \\
\cmidrule(lr){5-7}
\cmidrule(lr){8-10}
& Regime &
Examples &
Time &
R@1 & R@5 & R@10 &
R@1 & R@5 & R@10 \\
\midrule
Video--text
& Compute-matched
& $\sim$3.5K
& $<0.5$ h
& 43.3 & 66.6 & 75.5
& 35.8 & 60.9 & 71.8 \\

Video--text
& Full data
& $\sim$60K
& $\sim$10 h
& 51.8 & 73.6 & 81.9
& 51.4 & 74.8 & 81.4 \\

\midrule
Text only
& ours
& $\sim$60K
& $<0.5$ h
& \textbf{52.5} & \textbf{76.3} & \textbf{83.8}
& \textbf{53.7} & \textbf{79.4} & \textbf{85.0} \\
\bottomrule
\end{tabular}
}
\end{table}
\vspace{-1.5em}
Additional ablation studies on alternative text-to-text mapping strategies, optimization objectives, prompt sensitivity, and zero-shot intermediate-layer performance are provided in Appendix~\ref{app:ablations}.

\vspace{-3mm}
\section{In-context Optimization Analysis}
\label{sec:analysis}
\vspace{-3mm}

In this section, we examine our in-context optimization approach beyond retrieval performance.
\begin{figure}[b]
\centering
\vspace{-5mm}
\begin{minipage}[b]{0.4\linewidth}
\includegraphics[width=0.9\linewidth]{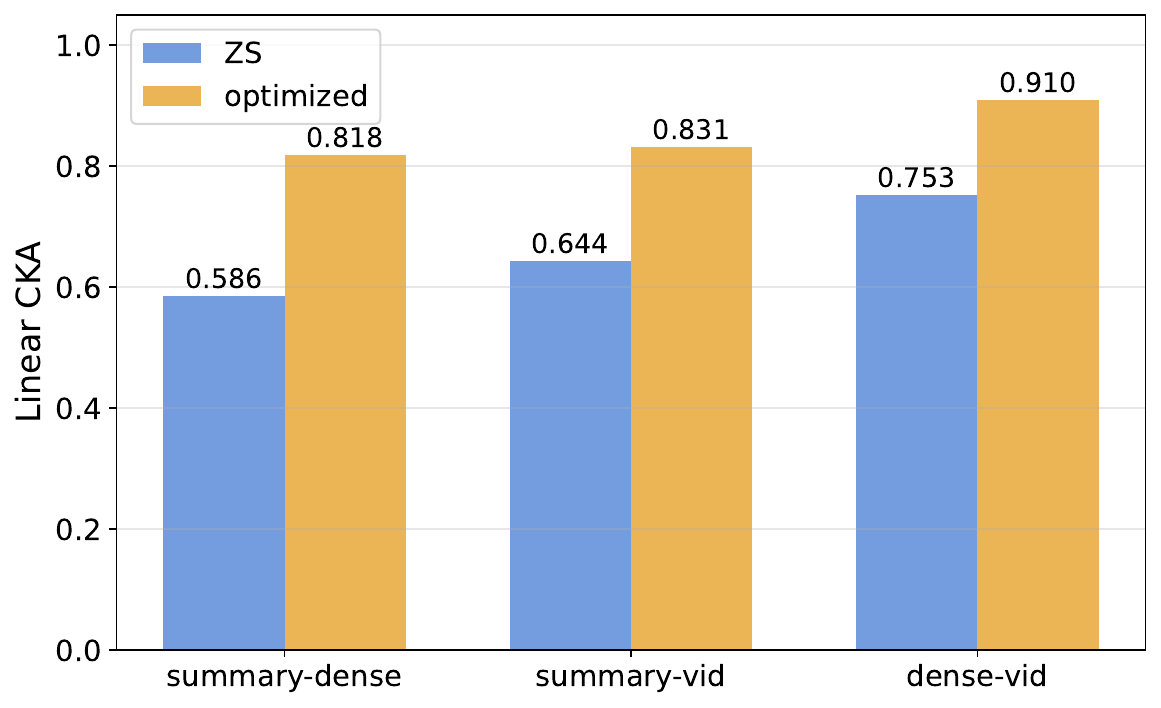}
\captionsetup{font=scriptsize}
\caption{CKA between video embeddings, dense-caption embeddings, and short-summary embeddings before and after our text-only dense-caption--summary optimization. Optimization improves text-text and text-visual alignment.
}
\label{fig:cka}
\end{minipage}
\hfil
\begin{minipage}[b]{0.4\linewidth}
\includegraphics[width=\linewidth]{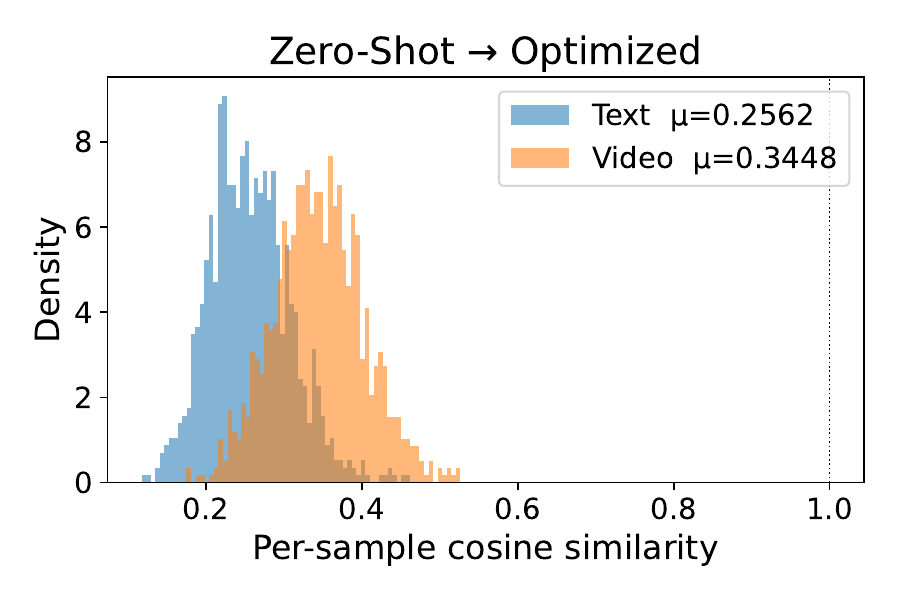}
\captionsetup{font=scriptsize}
\caption{Similarity between each sample and its corresponding pre-optimization representation. The low similarities indicate feature drift in both modalities.}
\label{fig:modality-drift}
\end{minipage}
\vspace{-5mm}
\end{figure}

\vspace{-2mm}
\paragraph{In-Context Textual Proxy.}
To examine the relationship between video embeddings, dense captions, and short summaries, we compute their Centered
Kernel Alignment (CKA) \cite{kornblith2019similarity} values before and after our [dense caption]-[short summary] optimization. CKA measures the similarity between two representation spaces by comparing their centered similarity/Gram matrices. It is useful for assessing whether two embedding sets preserve comparable structure, such as similar pairwise similarity patterns, being robust to orthogonal transformations and isotropic scaling, or feature dimension. Fig.~\ref{fig:cka} shows alignment patterns. In the zero-shot setting (blue), dense-caption--video alignment is the strongest among all pairwise alignments, suggesting that dense captions can serve as a proxy for the visual video representation in the MLLM embedding space. After optimization (yellow), all CKA values increase substantially. Despite the objective being defined only over [dense-caption]--[short-summary] pairs, we also obtain a significant increase in [short-summary]--[video] alignment which helps explain the improved retrieval performance, as short summaries better resemble textual queries. Moreover, although the optimization targets alignment in the text domain, the alignment between dense captions and videos is increased. Together, these results suggest that our optimization might not merely reshape the textual space, but also modifies an underlying latent video semantic space.

\vspace{-2mm}
\paragraph{In-context optimization changes both text and video representation spaces.}
To further analyze post-optimization change in the representation spaces, we measure each sample’s similarity to its pre-optimization representation and report the results in Fig.~\ref{fig:modality-drift}. The low similarity values indicate substantial post-optimization shifts in both modalities, suggesting that the optimization reshapes the shared multimodal geometry rather than merely improving text-pair alignment.

\vspace{-2mm}
\paragraph{Optimization promotes a more isotropic representation space for both modalities.}
Anisotropy is a well-documented property of neural embedding spaces, where representations concentrate in a narrow cone or along a few dominant directions, causing unrelated samples to exhibit high cosine similarity \cite{ethayarajh2019contextual,cai2021isotropy,mu2017all}. This issue also appears in modern vision--language models, whose shared embedding spaces can show modality-specific concentration, modality gaps, and limited uniformity \cite{liang2022mind,fahim2024s,eslami2025mitigate}.
Following \cite{ethayarajh2019contextual}, we quantify anisotropy $(\mathcal{A}\ \in [0,1])$ by the average cosine similarity between randomly sampled features (lower indicates more isotropic space).
In the zero-shot regime, both modalities exhibit strong anisotropy, with \(\mathcal{A}=0.84\) for text and \(0.86\) for video. After optimization, these values decrease to \(0.26\) and \(0.23\), respectively, indicating a substantially less collapsed representation space. A similar trend is observed for the inter-modal video-text space, where anisotropy drops from \(\mathcal{A}=0.78\) before optimization to \(\mathcal{A}=0.21\) afterward. These reductions suggest that optimization improves the cross-modal shared space structure.

\begin{figure}[t]
\vspace{-10mm}
\begin{minipage}[t]{0.45\linewidth}
\includegraphics[width=\linewidth]{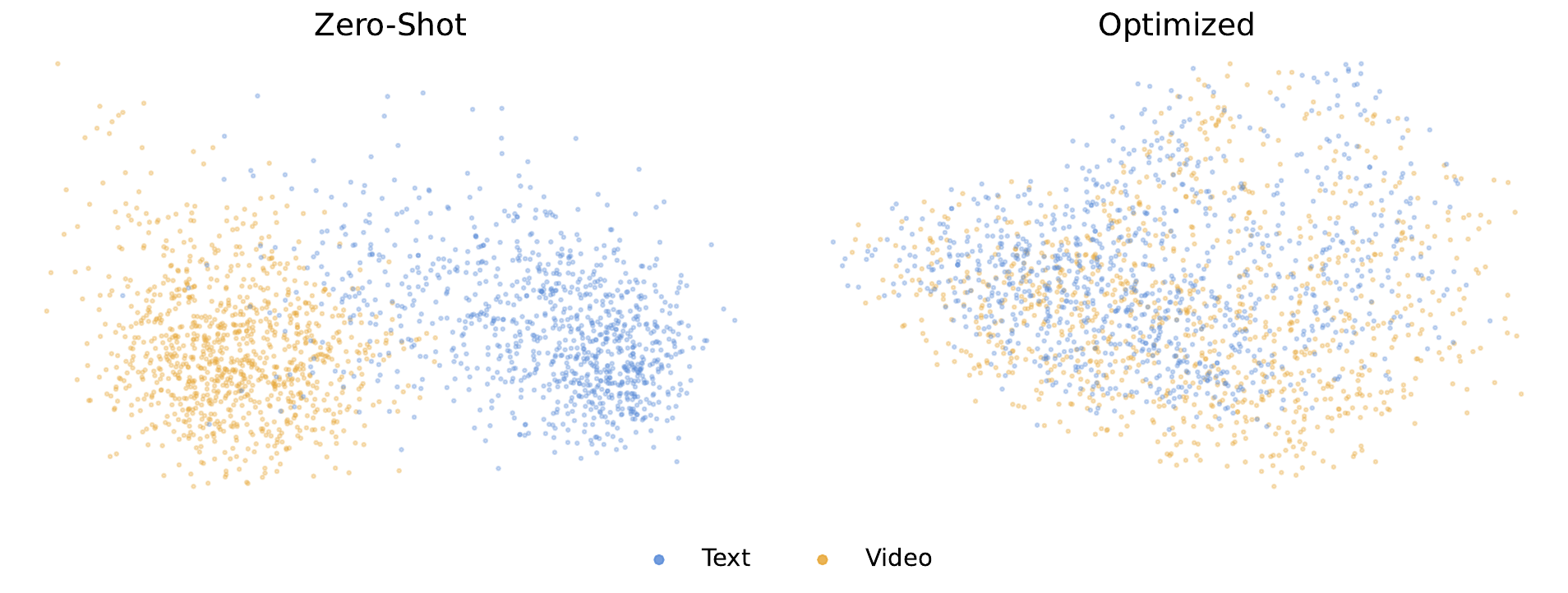}
\captionsetup{font=scriptsize}
\caption{Joint PCA visualization of text and video features before and after ourin-contexttext-only optimization.}\label{fig:pca}
\end{minipage}
\hfil
\begin{minipage}[t]{0.45\linewidth}
\includegraphics[width=\linewidth]{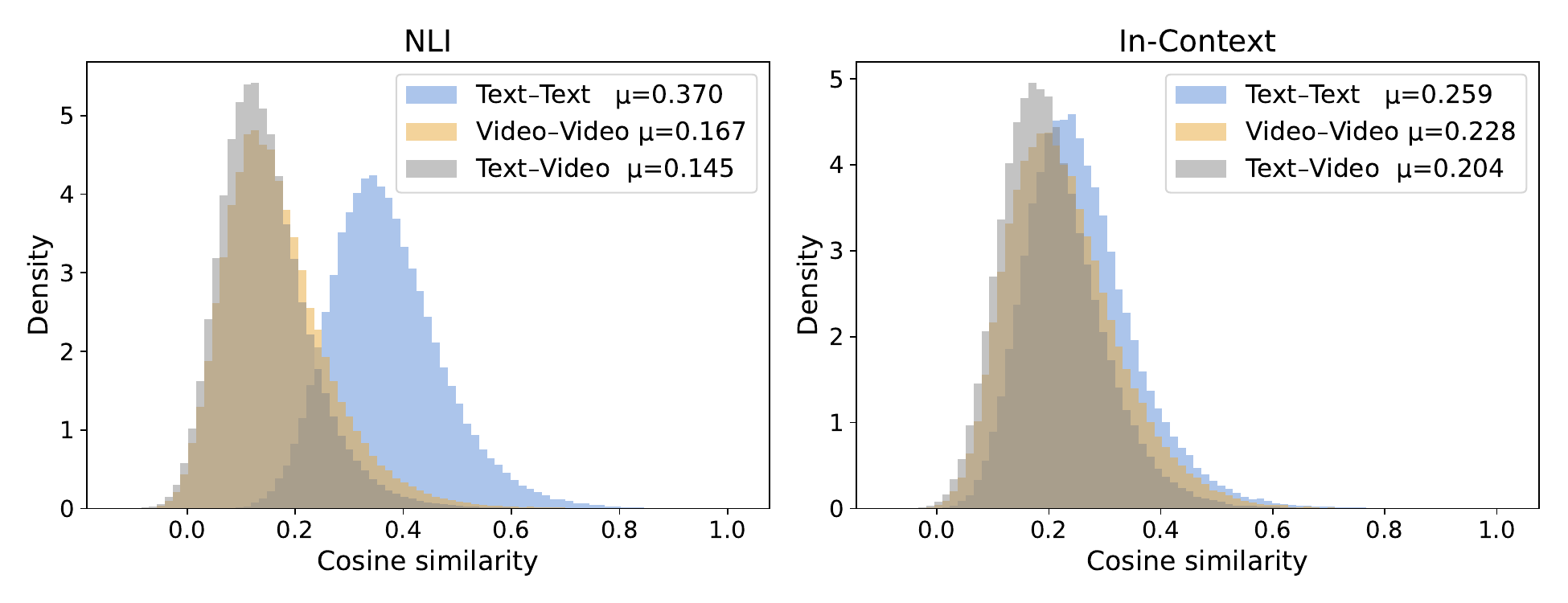}
\captionsetup{font=scriptsize}
\caption{Pairwise cosine similarity distributions for NLI- and in-context-optimized models, where in-context optimization yields more balanced intra- and cross-modal similarities.}
\label{fig:nli-vs-ours-dists}
\end{minipage}
\vspace{-5mm}
\end{figure}

\vspace{-3mm}
\paragraph{PCA visualization suggests a smaller modality gap.}
As a qualitative complement, we visualize text and video embeddings in a joint PCA subspace. In the zero-shot setting, the two modalities appear partially separated, suggesting a visible modality gap in the leading principal directions. After optimization the text and video embeddings appear more intermixed in a shared subspace. Thus, Fig.~\ref{fig:pca} qualitatively suggests that in-context optimization reduces the visible separation between modalities in the joint embedding space.

\vspace{-3mm}
\paragraph{In-context optimization is stronger than NLI for video--text alignment.}
Prior MLLM embedders use text-only NLI supervision \cite{E5-V_2024,LAMRA_CVPR25}, which provides a generic language-alignment signal rather than a video-grounded one. Fig.~\ref{fig:nli-vs-ours-dists} compares NLI with our in-context optimization on 500 random samples. NLI yields high Text--Text similarity, but much lower Video--Video and Text--Video similarities, indicating a text-dominant space. In contrast, our in-context objective, based on dense captions and short summaries of the same video, produces more balanced similarities across modalities, suggesting stronger video--text alignment and explaining its gains over NLI-based LamRA in Tab.~\ref{tab:united_t2v_retrieval},\ref{tab:united_v2t_retrieval}.

Overall, our analysis indicates that the zero-shot model already contains latent cross-modal structure, but this structure is partially obscured by anisotropy and modality gap. The proposed proxy objective reduces this mismatch without requiring direct video--text supervision.

\vspace{-3mm}
\section{Summary}
\vspace{-3mm}
In this work, we study off-the-shelf MLLMs for the task of video–text embedding and retrieval. We start by revealing a previously unrecognized retrieval capability (aligned embedding extraction) already embedded in video MLLMs without further training. To further improve multimodal alignment, we introduce an efficient in-context optimization scheme that relies solely on text-to-text mapping, requiring no visual supervision. When tested on video retrieval, the resulting embeddings significantly outperform recent trained MLLM embedders, and our complete approach achieves state-of-the-art results across multiple benchmarks.
We further demonstrate the efficiency and retrieval advantages of text-only optimization over its paired video--text-trained counterpart, and provide an in-depth analysis of its impact on both textual and visual domains, providing explainability for this effect.
More broadly, this points to pretrained MLLMs as not only generative models, but also data-efficient embedders whose retrieval capabilities can be exposed through carefully designed text-only supervision.

\bibliographystyle{plainnat}
\bibliography{reference}

\input{appendix}

\end{document}

%% file: appendix.tex
\clearpage
\appendix
\section*{Appendix}
\vspace{-3mm}
This Appendix includes details on the following topics:
\begin{enumerate}[itemsep=2pt, topsep=2pt, parsep=0pt, partopsep=0pt]
  \item[\textbf{A.}] Benchmark Datasets
  \item[\textbf{B.}] Generalization Across Backbones
  \item[\textbf{C.}] Controlled Video–Text Supervision
  \item[\textbf{D.}] Evaluations on more datasets
  \item[\textbf{E.}] Ablation Study
  \item[\textbf{F.}] Zero-Shot Study
  \item[\textbf{G.}] Reranking
  \item[\textbf{H.}] More implementation details
  \item[\textbf{I.}] Further In-Context Optimization Analysis
  \item[\textbf{J.}] Limitations
  \item[\textbf{K.}] Societal Impact
  \item[\textbf{L.}] Licenses
\end{enumerate}

\vspace{-3mm}
\section{Benchmark Datasets}
In this section we provide detailed information on the evaluated benchmarks - MSR-VTT, MSVD, VATEX, DiDeMo and ActivityNet.
A summary of the datasets is provided in Tab.~\ref{tab:datasets-summary}.

	\noindent
    \textbf{MSR-VTT} \cite{xu2016msrvtt} contains 10,000 videos (10–32s each) with 200,000 associated captions. We follow the standard 1k-A test split \cite{Webvid_ICCV2023, Clip4clip2021}, which consists of 1,000 video–text pairs.

    \noindent
    \textbf{MSVD} \cite{chen2011collecting_msvd} contains 1,970 videos (1–62s). The train/validation/test splits include 1,200/100/670 videos, respectively. Each video has approximately 40 English captions. In the V2T setting, we treat any caption for a given video as a positive.

	\noindent
	\textbf{VATEX} \cite{wang2019vatex} contains 25,991 training videos, 3,000 validation videos, and 6,000 test videos, with 10 paired captions per video in both English and Chinese. Following \cite{chen2020fine}, we evaluate on 1,500 videos from the validation set using English captions only.

    \noindent
    \textbf{DiDeMo} \cite{anne2017localizing_didemo} contains over 10,000 videos (25-30s), annotated with natural language descriptions of temporally localized events. Following \cite{Gabeur2020MMT, Clip4clip2021}, the descriptions associated with each video are concatenated into a single paragraph, and video–paragraph retrieval is evaluated on the standard test split, comprising 1,004 video–text pairs.

    \noindent
    \textbf{ActivityNet} \cite{caba2015activitynet,krishna2017dense_activitynet} contains 20,000 videos. Following \cite{Gabeur2020MMT, Clip4clip2021}, we concatenate all descriptions for each video into a single paragraph and evaluate video–paragraph retrieval on the `val1' split.

\begin{table}[h]
\centering
\caption{Benchmark datasets statistics, with standard test subsets.}
\label{tab:datasets}
\small
\setlength{\tabcolsep}{6pt}
\begin{tabular}{lcc}
\toprule
Dataset & \#Videos & \#Captions \\
\midrule
MSRVTT       & 1,000 & 1,000 \\
MSVD        & 670  & 27,763 \\
VATEX        & 1,500 & 15,000 \\
DiDeMo       & 1,004 & 1,004 \\
ActivityNet  & 4,917 & 4,917 \\
\bottomrule
\end{tabular}
\vspace{-3mm}
\label{tab:datasets-summary}
\end{table}

\section{Generalization Across Backbones}
\label{app:backbone_generalization}

To evaluate the generalizability of our proposed text-only optimization, we extend our experiments beyond the default VideoLLaMA3 backbone to two additional video MLLMs: Qwen3-VL and Keye-VL-1.5. For each backbone, we apply the same text-only in-context optimization procedure and evaluate the resulting representations under identical settings on four video-text retrieval benchmarks: MSR-VTT, MSVD, VATEX, and DiDeMo. Tables~\ref{tab:backbones_t2v} and~\ref{tab:backbones_v2t} report text-to-video (T2V) and video-to-text (V2T) retrieval performance, respectively.

The results demonstrate that our text-only optimization can be applied to multiple video MLLM backbones, yielding strong retrieval performance across all four benchmarks and both retrieval directions. In particular, Keye-VL-1.5 achieves performance comparable to VideoLLaMA3, occasionally surpassing it at R@10, while Qwen3-VL also exhibits competitive performance despite generally underperforming the other two backbones. These findings support the applicability of our approach beyond the default backbone, while indicating that retrieval performance remains dependent on the underlying model.

\begin{table*}[t]
\small
\centering
\captionsetup{font=small}
\caption{Text-to-video retrieval performance across different video MLLM backbones. All models are evaluated after applying our text-only optimization procedure. Results are reported as Recall@K.}
\label{tab:backbones_t2v}
\resizebox{\textwidth}{!}{\begin{tabular}{l|ccc|ccc|ccc|ccc}
\toprule
& \multicolumn{3}{c|}{MSR-VTT}
& \multicolumn{3}{c|}{MSVD}
& \multicolumn{3}{c|}{VATEX}
& \multicolumn{3}{c}{DiDeMo} \\
\cmidrule(lr){2-4}
\cmidrule(lr){5-7}
\cmidrule(lr){8-10}
\cmidrule(l){11-13}
Backbone
& R@1 & R@5 & R@10
& R@1 & R@5 & R@10
& R@1 & R@5 & R@10
& R@1 & R@5 & R@10 \\
\midrule
Qwen3-VL
& 50.1 & 73.2 & 82.6
& 52.6 & 79.2 & 86.3
& 60.4 & 89.6 & 94.8
& 46.9 & 72.0 & 79.1 \\
Keye-VL-1.5
& 51.7 & 75.7 & \textbf{84.4}
& 59.8 & 84.3 & 89.8
& 68.0 & 91.3 & 95.6
& 51.2 & 77.5 & 83.2 \\
\midrule
VideoLLaMA3
& \textbf{52.5} & \textbf{76.3} & 83.8
& \textbf{60.8} & \textbf{84.9} & \textbf{90.1}
& \textbf{68.2} & \textbf{93.6} & \textbf{96.8}
& \textbf{53.7} & \textbf{79.4} & \textbf{85.0} \\
\bottomrule
\end{tabular}}
\end{table*}

\begin{table*}[t]
\centering
\small
\captionsetup{font=small}
\caption{Video-to-text retrieval performance across different video MLLM backbones. All models are evaluated after applying our text-only optimization procedure. Results are reported as Recall@K.}
\label{tab:backbones_v2t}
\resizebox{\textwidth}{!}{\begin{tabular}{l|ccc|ccc|ccc|ccc}
\toprule
& \multicolumn{3}{c|}{MSR-VTT}
& \multicolumn{3}{c|}{MSVD}
& \multicolumn{3}{c|}{VATEX}
& \multicolumn{3}{c}{DiDeMo} \\
\cmidrule(lr){2-4}
\cmidrule(lr){5-7}
\cmidrule(lr){8-10}
\cmidrule(l){11-13}
Backbone
& R@1 & R@5 & R@10
& R@1 & R@5 & R@10
& R@1 & R@5 & R@10
& R@1 & R@5 & R@10 \\
\midrule
Qwen3-VL
& 51.5 & 75.8 & 83.0
& 77.5 & 89.6 & 93.1
& 84.3 & 96.8 & 98.3
& 48.0 & 71.4 & 79.1 \\
Keye-VL-1.5
& \textbf{54.9} & 77.2 & \textbf{85.2}
& 84.9 & \textbf{95.5} & \textbf{98.1}
& 88.9 & 97.1 & 98.9
& 51.9 & 75.6 & 81.8 \\
\midrule
VideoLLaMA3
& \textbf{54.9} & \textbf{77.5} & 84.1
& \textbf{85.7} & \textbf{95.5} & 97.8
& \textbf{89.6} & \textbf{98.5} & \textbf{99.3}
& \textbf{56.5} & \textbf{79.7} & \textbf{86.0} \\
\bottomrule
\end{tabular}}
\end{table*}

\section{Controlled Video--Text Supervision}
\label{app:video-text-supervision}

In Sec.~\ref{sec:video-text-supervision} of the main paper, we directly compare our text-only optimization with conventional video--text fine-tuning using a controlled baseline trained on the same VideoUFO examples. Specifically, we retrieve the corresponding video clips for the $\sim$60K examples used in text-only optimization and replace the dense-caption--summary supervision with video--brief-caption pairs.

We keep the VideoLLaMA3 backbone, LoRA configuration, optimization objective, data split, and training recipe fixed. Since processing raw video is substantially more expensive than processing text, we evaluate the video--text baseline under two settings: (i) a compute-matched setting, in which the number of training examples is reduced to approximately match the training compute of our text-only method, and (ii) a full-data setting, in which the video--text baseline is trained on all $\sim$60K examples.

As shown in Tab.~\ref{tab:video-text-supervision}, text-only optimization is substantially more compute-efficient than direct video--text training. Under a matched training budget, the text-only approach can use substantially more training examples from distinct videos and consequently achieves considerably stronger retrieval performance. For example, on DiDeMo, R@1 increases from 35.8 to 53.7.

Interestingly, the advantage is not explained by compute efficiency alone. Even when the video--text baseline is trained on all approximately 60K examples, requiring roughly $20\times$ more training time, it remains below the text-only model across all reported metrics. One possible explanation is that dense captions provide a compact and semantically explicit representation of the video content relevant to text alignment, whereas raw videos contain substantial temporal redundancy and require processing many visual tokens. Consequently, learning the same semantic alignment from video may be both computationally more expensive and more difficult to optimize.

These results demonstrate that, in the controlled VideoLLaMA3 setting, dense-caption--summary supervision is both more computationally efficient and more effective than direct video--text fine-tuning. However, this finding does not necessarily generalize to other architectures, objectives, or training regimes.

\newpage
\section{More Evaluations}
\vspace{-2mm}
In this section, we report additional results for \ours.
\vspace{-2mm}

\subsection{MMEB-v2 Comparison}
\vspace{-2mm}
Recently, \cite{VLM2Vec-V2_2025} suggested a new benchmark, MMEB-v2, for video tasks, which includes few text-to-video retrieval datasets. In Tab.~\ref{tab:MMEBv2} we compare our method with existing MLLM embedders, using the benchmark’s publicly available results. We report results on MSR-VTT and DiDeMo, whose text and video splits follow the standard evaluation and were not modified in MMEB-v2. \ours raw scores outperform other methods also here. However, note that the LamRA paper reports substantially higher results (including for VLM2Vec), suggesting that the MMEB-v2 evaluation protocol or implementation may be suboptimal.

\vspace{-3mm}
\begin{table}[H]
\centering
\small
\captionsetup{font=small}
\caption{Comparison to MMEB-v2 results on Text-to-Video retrieval (Recall@K).}
\scalebox{0.9}{
\begin{tabular}{l|ccc|ccc}
\toprule
 & \multicolumn{3}{c|}{MSR-VTT} & \multicolumn{3}{c}{DiDeMo} \\

Model & R@1 & R@5 & R@10 & R@1 & R@5 & R@10 \\
\midrule
LamRA       & 25.0 & 47.4 & 57.6 & 22.8 & 44.7 & 56.9 \\
VLM2Vec     & 34.5 & 58.5 & 67.9 & 29.3 & 53.5 & 62.2 \\
VLM2Vec-v2  & 28.3 & 50.4 & 59.4 & 30.4 & 53.9 & 63.1 \\
UniME-v2    & 27.6 & 44.9 & 52.9 & 31.5 & 53.0 & 61.9 \\
\midrule
\ours-O      & \textbf{48.9} & \textbf{71.4} & \textbf{81.1} & \textbf{43.0} & \textbf{69.0} & \textbf{76.3} \\
\bottomrule
\end{tabular}
}
\label{tab:MMEBv2}
\end{table}

\vspace{-3mm}
\subsection{ActivityNet Evaluation}
\vspace{-2mm}
We conduct an additional evaluation on ActivityNet~\cite{caba2015activitynet}. Tab.~\ref{tab:activitynet_t2v_v2t_retrieval} summarizes the results on this dataset. Since VideoLLaMA3 was exposed to ActivityNet data during its generative training, we exclude these results from the main paper; nevertheless, we report them here for completeness and to support future studies. Notably, \ours outperforms even methods that were directly finetuned on ActivityNet's training set (+5.1\% gain relative to a fine-tuned InternVideo2-6B)

\vspace{-3mm}
\begin{table}[H]
\centering
\small
\captionsetup{font=small,width=0.8\linewidth}
\caption{ActivityNet retrieval (R@K): Text-to-Video (left) and Video-to-Text (right). \emph{Finetuned on ActivityNet:} CLIP4Clip (ft), ViCLIP (ft), UMT-L (ft), InternVideo2-6B (ft).}
\scalebox{0.93}{
\begin{tabular}{l|ccc|ccc}
\toprule
& \multicolumn{3}{c|}{Text-to-Video (T2V)} & \multicolumn{3}{c}{Video-to-Text (V2T)} \\
Model & R@1 & R@5 & R@10 & R@1 & R@5 & R@10 \\
\midrule
\multicolumn{7}{l}{\it{Multimodal LLM Embedders}} \\
LamRA      & 58.5 & 85.4 & 91.6 & 58.7 & 84.8 & 92.5 \\
VLM2Vec    & 32.8 & 52.9 & 61.3 & 40.4 & 59.2 & 66.2 \\
MMRet-v1.5 & 50.6 & 76.6 & 85.8 & 52.7 & 77.8 & 86.8 \\
B3         & 33.8 & 51.9 & 59.6 & 39.2 & 57.4 & 65.3 \\
VLM2Vec-V2 & 35.5 & 54.4 & 61.4 & 38.8 & 58.0 & 65.6 \\
UNITE      & 25.8 & 42.9 & 51.6 & 31.4 & 47.6 & 55.0 \\
UniME-V2   & 31.4 & 50.9 & 59.9 & 38.5 & 57.2 & 64.5 \\
\midrule
\multicolumn{7}{l}{\it{Video Foundation Models}} \\
ViCLIP           & 15.1 & -    & -    & 24.0 & -    & -    \\
InternVideo-L    & 30.7 & -    & -    & 31.4 & -    & -    \\
UMT-L            & 42.8 & 69.6 & 79.8 & 40.7 & 67.6 & 78.6 \\
LanguageBind     & 41.0 & 68.4 & 80.0 & -    & -    & -    \\
VideoPrism-g     & 52.7 & 79.4 & -    & 50.3 & 77.1 & -    \\
PE-Core-G        & 54.7 & -    & -    & 51.2 & -    & -    \\
InternVideo2-6B  & 63.2 & 85.6 & 92.5 & 56.5 & 82.8 & 90.3 \\
\midrule
\multicolumn{7}{l}{\it{Fine-tuned on ActivityNet}} \\
CLIP4Clip (ft)   & 40.3 & -- & -- & 41.6 & -- & -- \\
ViCLIP (ft)      & 49.8 & -- & -- & 48.1 & -- & -- \\
UMT-L (ft)       & 66.8 & -- & -- & 64.4 & -- & -- \\
InternVideo2-6B (ft) & 74.1 & -- & -- & 69.7 & -- & -- \\
\midrule
\ours & \textbf{78.4} & \textbf{89.7}  & \textbf{95.2}  & \textbf{71.9} & \textbf{90.8}  &  \textbf{95.8} \\
\bottomrule
\end{tabular}}
\label{tab:activitynet_t2v_v2t_retrieval}
\end{table}
\vspace{-3mm}

In this context, we also revisit Qwen3-VL-Embedding, for which non-exposure to benchmark training sets is not explicitly stated. Its results are comparable to the \emph{fine-tuned} regime, where models are trained on the corresponding benchmark splits \cite{Clip4clip2021,InternVideo2022,Internvideo2_ECCV2024}. Specifically, on MSR-VTT, VATEX, and ActivityNet, Qwen3-VL-Embedding achieves T2V R@1 scores of 61.3, 71.6, and 75.3, respectively, compared with 62.8, 75.5, and 74.1 for fine-tuned InternVideo2-6B.

\section{Ablation Study \protect\footnote{The ablation experiments through the appendix were conducted without dual-softmax to isolate the effect of the examined component.}}
\label{app:ablations}

\subsection{Impact of In-Context Optimization}

\begin{figure}[H]
\centering
\begin{minipage}[t]{0.25\columnwidth}
\vspace{15pt}
\centering
\small
\captionsetup{font=small}
\captionof{table}{Effect of different textual pairs on MSR-VTT T2V Recall@1.}
\label{tab:ablation_textual_choice}
\begin{tabular}{lc}
\toprule
Context & R@1 \\
\midrule
Zero-shot       & 14.3 \\
NLI (LamRA)             & 45.0 \\
Video Captions  & 46.7 \\
LVD-2M (YT)         & 47.2 \\
In-Context & 47.7 \\
\bottomrule
\end{tabular}
\end{minipage}
\hfill
\begin{minipage}[t]{0.25\columnwidth}
\vspace{15pt}
\centering
\small
\captionsetup{font=small}
\captionof{table}{Effect of optimization Loss on MSR-VTT T2V Recall@1.}
\label{tab:ablation_optimization_effect}
\begin{tabular}{lc}
\toprule
Optimization & \multirow{2}{*}{R@1} \\
Strategy     &                      \\
\midrule
Zero-shot    & 14.3 \\
InfoNCE      & 47.7 \\
DSL          & 48.9 \\
\bottomrule
\end{tabular}
\end{minipage}
\hfill
\begin{minipage}[t]{0.45\columnwidth}
\vspace{0pt}
\centering
\includegraphics[width=0.9\linewidth]{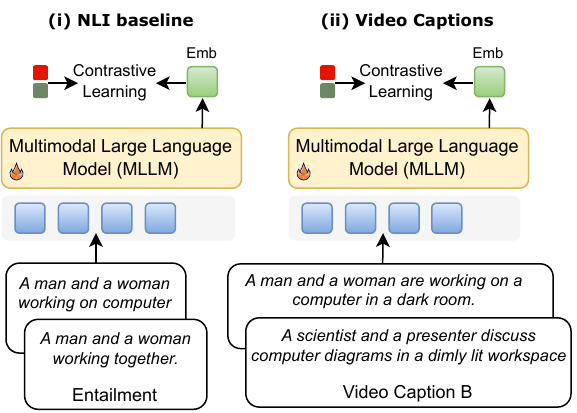}
\captionsetup{font=small}
\captionof{figure}{Illustration of the different textual data variants used in our ablation study.}
\label{fig:in-context-review}
\end{minipage}

\end{figure}

We evaluate the impact of different textual optimization approaches on Recall@1 performance on MSR-VTT. To distinguish the benefits of video-related textual supervision from those of our proposed in-context formulation, we compare against optimization using brief video description pairs (\textit{Video Captions}), in addition to the NLI-based textual data used in prior work. An illustration of NLI and \textit{Video Captions} is provided in Fig.~\ref{fig:in-context-review}, while our in-context approach is described in Fig.~\ref{fig:main}.

For \textit{Video Captions} optimization, we use the same data split as our in-context optimization. We prompt an LLM (Gemma3 \cite{gemma3}) to generate a revised short caption from the dense caption, avoiding the original short caption. This approach improves over NLI-based optimization, demonstrating the benefit of video-related textual supervision, while our in-context approach performs higher (46.7\% vs. 47.7\% R@1).

To further assess the impact of caption source, we repeat the optimization using dense--brief caption pairs from the YouTube-clips subset of LVD-2M \cite{xiong2024lvd}, while keeping the model and training protocol unchanged. This yields comparable, albeit slightly lower, performance (46.2\% vs. 46.7\% R@1), suggesting that the benefits generalize across caption datasets, although the choice of caption source can influence retrieval performance.

\subsection{Optimization Loss}
In Table \ref{tab:ablation_optimization_effect} we further report the impact of Dual Softmax Loss (DSL) (only in train) on the results.

\subsection{Prompt Selection}
We ablate several prompts for embedding extraction, all following the EOL prompting scheme, and find that task-aware prompting plays an important role in off-the-shelf MLLM video processing. In particular, adding the prefix prompt \texttt{``Recover the main subject or subjects, appearance and setting, and main activity in the video''} leads to a noticeable performance improvement. We therefore adopt this prefix for the main zero-shot video--text retrieval experiments.
After optimization, however, the ablated prompts yield results comparable to the standard prompt used in prior work, \texttt{``Summarize above sentence/video in one word: ''}. To ensure a clean comparison, we use this standard prompt for both the optimized setting and the remaining zero-shot experiments.
\newpage
\section{Zero-Shot Study}
\label{sec:zs-study}
\vspace{-2mm}
We study how zero-shot retrieval performance varies across the internal layers of off-the-shelf MLLMs. Specifically, we ablate the choice of layer used for embedding extraction across multiple Qwen-VL generations and selected video MLLMs of comparable scale, chosen based on their performance on the Video-MME short-clip benchmark \cite{fu2025videomme}. We evaluate Qwen2-VL, Qwen2.5-VL-7B, Qwen3-VL-8B, VideoLLaMA3-7B, and Keye-VL-8B. For each backbone, representations are extracted from the selected layer using the embedding extraction procedure described in Sec.~\ref{sec:feature_extraction}, and performance is measured on MSR-VTT text-to-video retrieval using Recall@1.

\begin{figure}[H]
  \centering
  \begin{minipage}[t]{0.44\textwidth}
  \vspace{0pt}
  \begin{tikzpicture}
  \node[anchor=south west, inner sep=0] (img) at (0,0)
{\includegraphics[width=\linewidth]{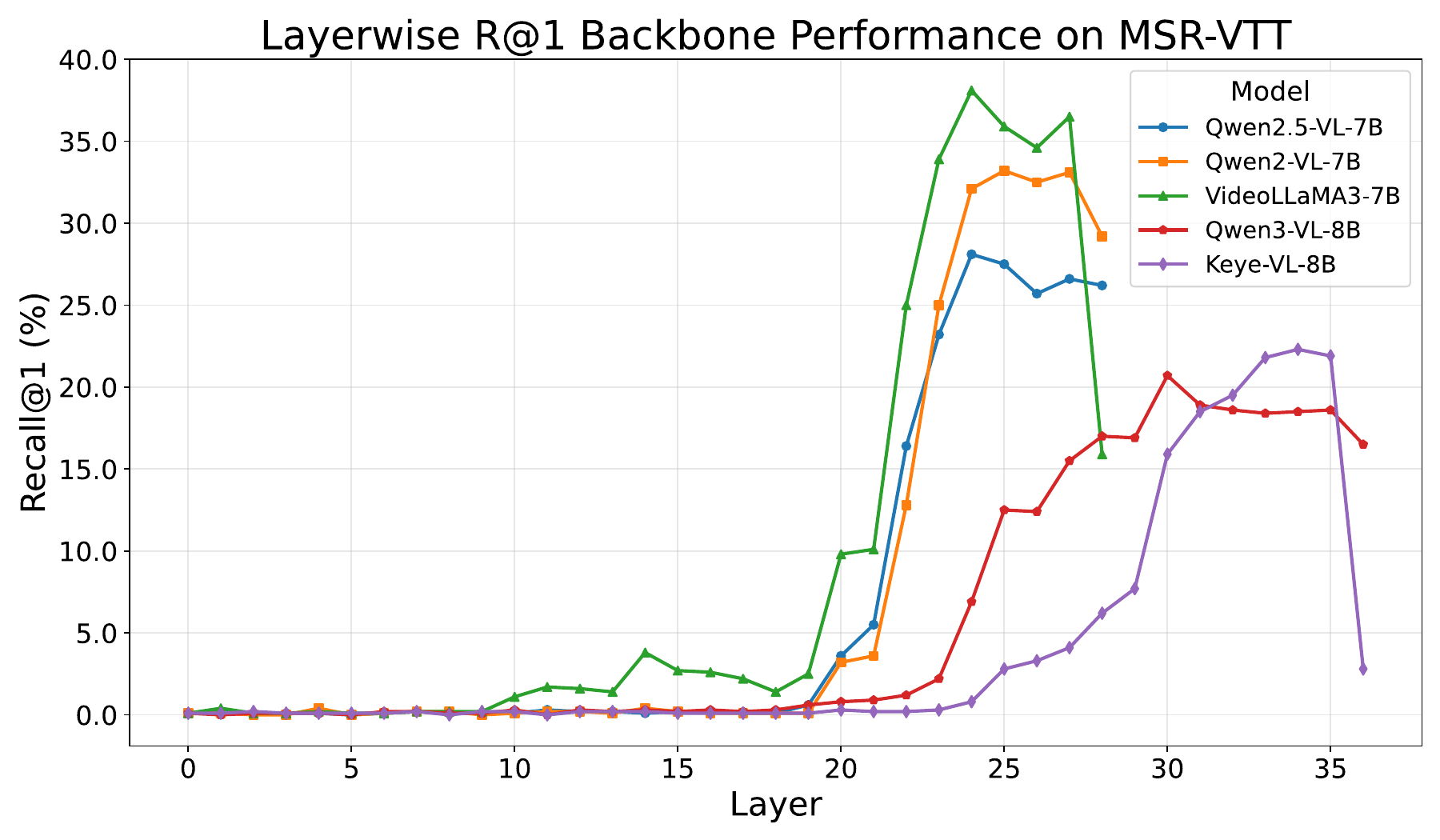}};
  \begin{scope}[x={(img.south east)}, y={(img.north west)}]
    \fill[white] (0.10,0.93) rectangle (0.90,1.0);
  \end{scope}
\end{tikzpicture}
    \captionsetup{font=scriptsize}
      \caption{Layer-wise zero-shot embedding performance across different MLLMs on MSR-VTT text-to-video Recall@1. While deeper layers generally yield stronger retrieval performance, the optimal results are not achieved at the final layer. VideoLLaMA3-7B attains the best overall performance.}

    \label{fig:layerwise-analysis}
  \end{minipage}
\hfill
  \begin{minipage}[t]{0.53\textwidth}
  \vspace{0pt}
  \centering
  \begingroup\setlength{\unitlength}{\textwidth}\begin{picture}(1,0.48833517)\put(0,0){\includegraphics[width=\unitlength,page=1]{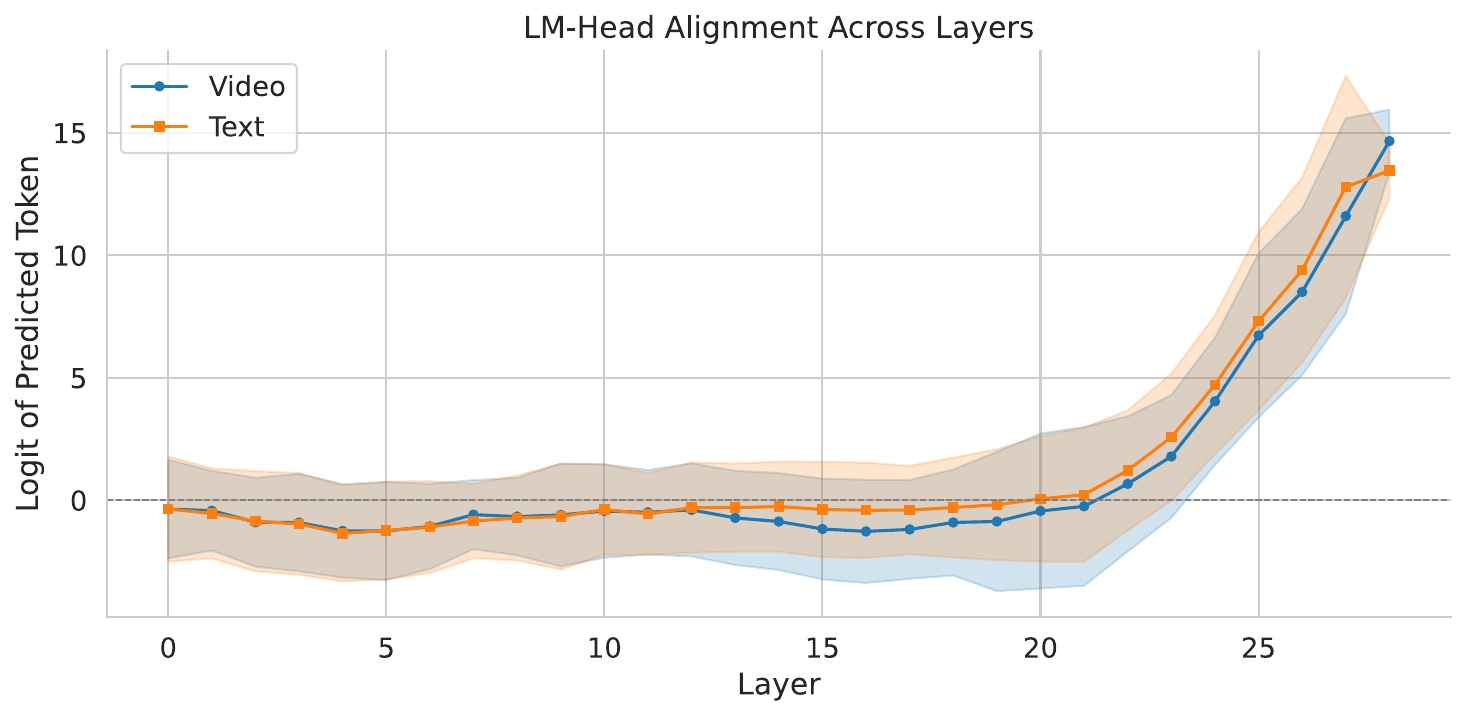}}\end{picture}\endgroup
  \captionsetup{font=scriptsize}
    \caption{Layer-wise probing with the LM head on VideoLLaMA3 in the zero-shot MSR-VTT setting. Higher layers are increasingly shaped by the LM head's next-token prediction objective.}
    \label{fig:layerwise-lm-head}
\vspace*{\fill}
    \end{minipage}
\end{figure}

As shown in Fig.~\ref{fig:layerwise-analysis}, layer selection has a substantial impact on zero-shot retrieval performance~\cite{layer_by_layer_ICML2025, effovpr_ICLR2025}. Early layers contain little retrieval-relevant signal, whereas several mid- to late-stage layers achieve markedly stronger performance, even without explicit alignment or retrieval-specific fine-tuning. Although deeper layers generally improve retrieval accuracy, the final layer is not optimal in the zero-shot setting. Across the evaluated backbones, VideoLLaMA3-7B achieves the strongest overall performance, with layer 24 yielding the best Recall@1. We therefore use VideoLLaMA3-7B as the default backbone, including for our lightweight in-context optimization, and use its layer-24 representations in the main zero-shot experiments.
\vspace{-3mm}
\vspace{-5mm}

\begin{figure}[H]
  \centering
  \adjustbox{trim=0 0 0 5mm,clip}{    \begingroup\setlength{\unitlength}{0.75\textwidth}\begin{picture}(1,0.40356396)\put(0,0){\includegraphics[width=\unitlength,page=1]{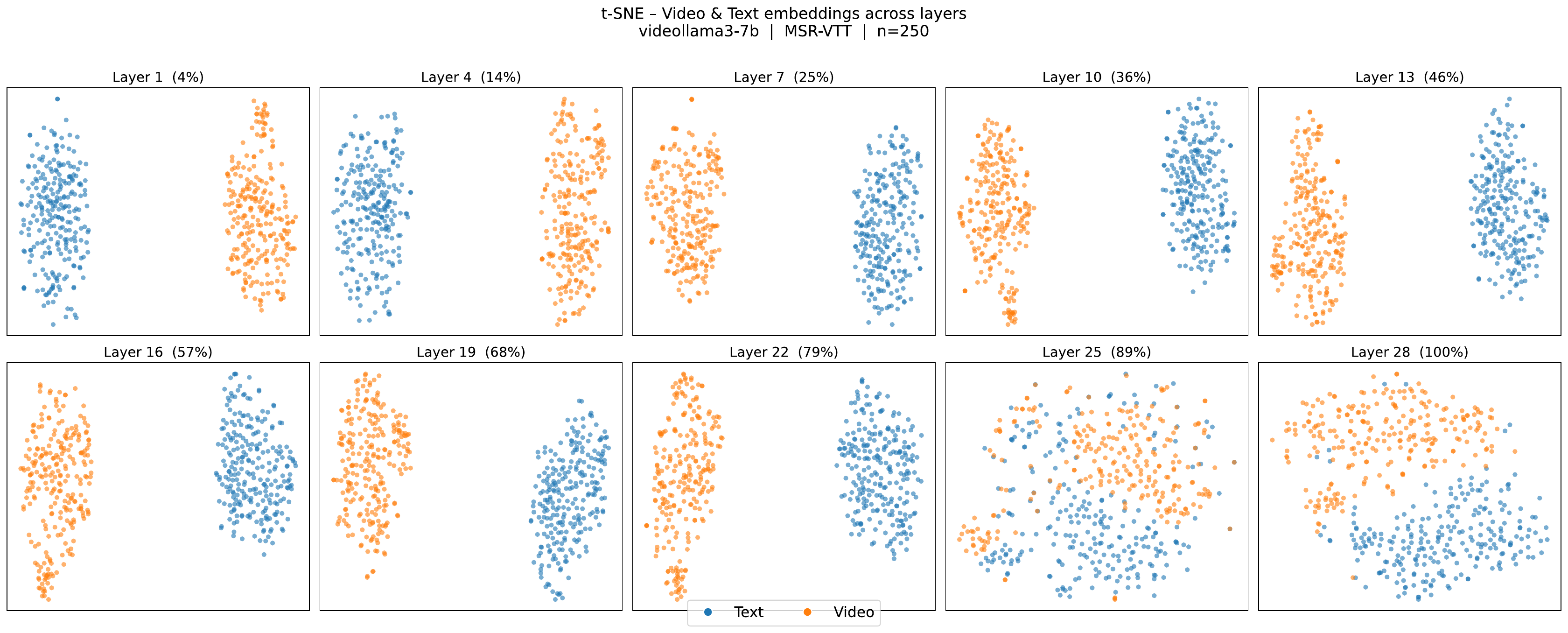}}\end{picture}\endgroup\   }
  \captionsetup{font=small,width=0.95\linewidth}
  \caption{Layer-wise t-SNE visualizations. In the early to intermediate layers, the embeddings of the two modalities occupy relatively distinct regions, indicating the presence of a modality gap. As the layers progress, however, the model eventually maps both modalities into a more closely aligned semantic space.}
  \label{fig:layerwise-tsne}
\end{figure}

To better understand why retrieval performance peaks at intermediate layers and degrades near the output layers, we conduct two complementary analyses. First, in Fig.~\ref{fig:layerwise-lm-head} we perform layer-wise probing using the model's next-token prediction layer, i.e., the LM head. This analysis shows that the final layers become increasingly coupled to the autoregressive generation objective. Second, in Fig.~\ref{fig:layerwise-tsne} we visualize representations across layers using t-SNE, which provides qualitative evidence of a depth-wise representational transition: early layers exhibit a pronounced modality gap, late-intermediate layers show stronger cross-modal semantic alignment. This pattern suggests that intermediate layers are most suitable for zero-shot retrieval because they encode shared video--text semantics while remaining less specialized to next-token prediction.
\vspace{-3mm}
\vspace{-3mm}
Interestingly, this behavior changes after optimization. In the optimized setting, the final-layer output becomes competitive with, and in practice preferable to, intermediate-layer embeddings: using the best-performing zero-shot intermediate layer after optimization yields only a negligible Recall@1 gain of less than one point. This indicates that optimization reduces the need for careful intermediate-layer selection, whereas such selection is crucial in the purely zero-shot setting.

\vspace{-3mm}
\subsection{Effective-Rank-Based Layer Selection}
\label{sec:erank_layer_selection}
\vspace{-2mm}
To provide a principled criterion for selecting a layer in a new backbone, we investigate the effective rank of the embedding Gram matrix (eRank) \cite{roy2007effective}. For each layer, we construct the Gram matrix of its embeddings, normalize its eigenvalues to sum to one, and compute eRank as the exponential of the entropy of the resulting eigenvalue distribution. We compute this score independently for the video and text modalities, average the two scores, and select the layer with the highest average eRank.

We evaluate this criterion on the MSR-VTT validation set across the five zero-shot MLLMs considered in this work. For four of the five backbones, eRank selects exactly the same layer as the retrieval-based sweep. Across all five models, the average performance gap relative to the oracle layer selected by the retrieval sweep is only 0.2 R@1 points, suggesting that eRank provides an effective retrieval-independent criterion for layer selection.

\vspace{-2mm}
\subsection{Computational Cost of Layer Selection}
\label{cost-layer-selecton}
\vspace{-2mm}

The layer-wise sweep is performed only once for each backbone, prior to training. After selecting the optimal layer, inference proceeds normally and incurs no additional overhead. The selection procedure is computationally inexpensive. We extract the hidden states of all intermediate layers in a single forward pass, rather than running the backbone separately for each candidate layer. The subsequent computation of the layer-selection metric introduces negligible additional cost.

We additionally find that the layer selected using the MSR-VTT validation set, containing 497 text-video samples, generalizes well across all evaluated datasets, avoiding the need to repeat the sweep for each dataset.

For our selected backbone, the complete layer-selection procedure takes approximately 1 minute and 45 seconds on a single NVIDIA B200 GPU with a batch size of 1, including model loading and setup. The encoding-time breakdown is reported in Tab.~\ref{tab:layer_selection_runtime}.

The additional computational cost of eRank is negligible. It reuses the hidden states extracted during the same forward pass used for layer evaluation and requires only one eigendecomposition per layer and modality. In our implementation, computing the eRank scores across all candidate layers takes less than one second.

\vspace{-3mm}
\begin{table}[H]
    \small
    \centering
    \captionsetup{font=small, width=0.87\linewidth}
    \caption{Encoding runtime for layer selection on the MSR-VTT validation set using the selected backbone. Hidden states from all intermediate layers are extracted in a single forward pass.}
    \label{tab:layer_selection_runtime}
    \begin{tabular}{lcc}
        \toprule
        Phase & Duration & Time per Sample \\
        \midrule
        Video encoding & 1 min 6 sec & $\sim$0.134 sec \\
        Text encoding  & 10 sec      & $\sim$0.021 sec \\
        \bottomrule
    \end{tabular}
\end{table}
\vspace{-3mm}
\vspace{-2mm}
\section{Reranking}
\vspace{-2mm}
\subsection{Effect of Reranking Depth}
\label{sec:reranking_depth}
\vspace{-2mm}
We further analyze the effect of the reranking depth $K$ on both retrieval performance and inference cost. Tab.~\ref{tab:reranking_depth} reports the effect of varying $K$ on MSR-VTT text-to-video retrieval.

Reranking consistently improves R@1 over the first-stage retrieval baseline. The best R@1 is obtained with $K=5$, while increasing the reranking depth leads to a small reduction in R@1. In contrast, larger values of $K$ can improve higher-rank recall, with $K=10$ yielding the best R@5 and $K=20$ the best R@10. These results indicate that shallow reranking captures most of the benefit, while larger reranking depths introduce additional inference cost with limited gains in performance.
\vspace{-3mm}
\begin{table}[H]
    \small
    \centering
    \captionsetup{font=small}
    \caption{Effect of reranking depth $K$ on MSR-VTT text-to-video retrieval.}
    \label{tab:reranking_depth}
    \begin{tabular}{lccc}
        \toprule
        Setting & R@1 & R@5 & R@10 \\
        \midrule
        First stage & 52.5 & 76.3 & 83.8 \\
        $K=5$  & 56.4 & 76.3 & 83.8 \\
        $K=10$ & 56.2 & 77.4 & 83.8 \\
        $K=20$ & 55.3 & 76.9 & 85.0 \\
        \bottomrule
    \end{tabular}
\end{table}
\vspace{-3mm}
The reranker requires 121 ms per query-candidate pair on a single NVIDIA B200 GPU. Since candidates are evaluated independently, sequential reranking therefore requires approximately 0.61 s, 1.21 s, and 2.42 s per query for $K=5$, $K=10$, and $K=20$, respectively.

\vspace{-3mm}
\subsection{Comparison with Reranking Methods}
\label{sec:reranker_comparison}
\vspace{-3mm}
We compare our calibrated-head reranker with two recent reranking approaches, RankVideo~\cite{skow2026rankvideo} and Qwen3-VL-Reranker~\cite{qwen3vlembedding}, while keeping the first-stage retriever and the top-$5$ candidate set fixed. Results on MSR-VTT text-to-video retrieval are reported in Tab.~\ref{tab:reranker_comparison}.
\vspace{-3mm}
\begin{table}[H]
    \small
    \centering
    \captionsetup{font=small, width=0.75\linewidth}
    \caption{Comparison of reranking methods on MSR-VTT text-to-video retrieval. All methods rerank the same top-$5$ candidates produced by the first-stage retriever.}
    \label{tab:reranker_comparison}
    \begin{tabular}{lc}
        \toprule
        Method & R@1 \\
        \midrule
        First stage & 52.5 \\
        RankVideo & 51.7 \\
        Qwen3-VL-Reranker & 57.1 \\
        Ours & 56.4 \\
        \bottomrule
    \end{tabular}
\end{table}
\vspace{-3mm}
RankVideo does not improve over the first-stage retriever in this setting, whereas both Qwen3-VL-Reranker and our calibrated-head reranker yield substantial gains. Qwen3-VL-Reranker achieves the highest R@1, outperforming our method by 0.7 points, suggesting that more sophisticated reranking approaches may offer additional improvements. Nevertheless, our lightweight approach remains competitive while requiring neither a dedicated reranking architecture nor reranker-specific training.

\vspace{-3mm}
\section{More Implementation Details}
\vspace{-3mm}
In Tab.~\ref{tab:mllm-embedders} we report full model names of MLLM embedders used in our evaluation.
\vspace{-3mm}
\begin{table}[H]
\centering
\caption{MLLM embedder baselines used in our evaluation.}
\small
\scalebox{0.9}{
\begin{tabular}{ll}
\toprule
Method & Model Name \\
\midrule
LamRA & LamRA-Ret \\
VLM2Vec & VLM2Vec-Qwen2VL-7B \\
MMRet-v1.5 & BGE-VL-v1.5-mmeb \\
B3 & B3\_Qwen2\_7B \\
VLM2Vec-V2 & VLM2Vec-V2.0 \\
UNITE & Unite-Base-Qwen2-VL-7B \\
UniME-V2 & UniME-V2-Qwen2VL-7B \\
\bottomrule
\end{tabular}
}
\label{tab:mllm-embedders}
\end{table}
\vspace{-3mm}

\vspace{-3mm}
\section{Further In-Context Optimization Analysis}
\label{sec:further-in-context}
\begin{figure}[H]
\centering
\includegraphics[width=0.8\linewidth]{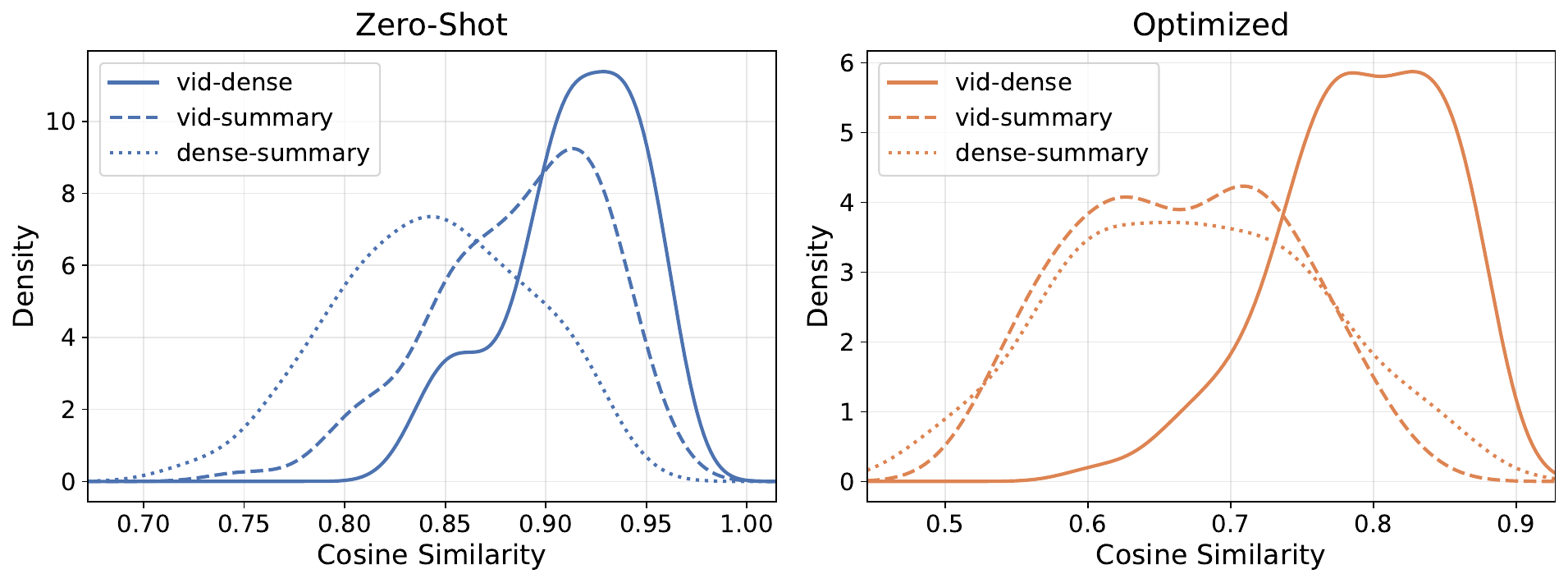}
\captionsetup{font=small,width=0.9\linewidth}
\caption{Pairwise similarity distributions among video embeddings, dense captions, and short summaries in the zero-shot setting and after in-context optimization. The high video--dense-caption similarity in the zero-shot setting indicates that dense captions can serve as a proxy for the visual video representation, while the post-optimization distribution shifts suggest improved alignment in a shared video--text semantic space. Results are computed on 100 triplets from the VideoUFO dataset.}
\label{fig:dist-pos-pairs}
\end{figure}

\vspace{-3mm}
To further understand our in-context optimization scheme, beyond the CKA results reported in the main paper, we examine the relationship between video embeddings, dense captions, and short summaries through pairwise similarity distributions. Overall, these distributions are consistent with the CKA findings. As shown in Fig.~\ref{fig:dist-pos-pairs}, in the zero-shot setting (blue), the higher video--dense-caption similarity suggests that dense captions can serve as a proxy for the visual video representation in the MLLM embedding space.
\vspace{-3mm}
\vspace{-3mm}
After optimization (Fig.~\ref{fig:dist-pos-pairs}, yellow), the video--summary similarity distribution closely follows the optimized dense-caption--summary distribution, suggesting that the learned text-only alignment induces a compatible structure in the video--text embedding space. Moreover, despite being optimized only toward short summaries, dense captions remain highly aligned with the corresponding video embeddings. Since video embeddings are never used in the objective, this indicates that the text-only optimization does not merely reshape the textual space, but also reinforces a shared latent semantic structure already present in the MLLM.

The overall post-optimization shift toward lower cosine similarities reflects an increased isotropy in the embedding space, as discussed in the main paper.

\textbf{In-Context Optimization.} We use LoRA by PEFT on the LLM backbone, optimizing its output token at the \texttt{<emb-1>} position. LoRA rank is $64$, and alpha is $128$. We run our single epoch optimization using deepspeed zero3, with 72 pairs per B200 GPU X 4 resulting in 288 in a batch.

\textbf{Evaluation.} Recent state-of-the-art VFMs (e.g., InternVideo2 and PE-Core) use dual-softmax score calibration, which accounts for the query distribution at inference time. To ensure a fair comparison, we apply the same protocol to our method and to all MLLM embedders. Specifically, we tune one fixed temperature per retrieval direction of T2V and V2T on the MSR-VTT validation set, scale the similarity matrix, and apply dual-softmax by taking the softmax over both columns and rows.

\textbf{Analysis}. The analysis experiments were done on MSR-VTT dataset, except when explicitly stated differently. For zero-shot setting, we used standard EOL prompt.

\section{Limitations}

\subsection{Textual Supervision}

Although our text-only optimization slightly outperforms its video--text-trained counterpart in overall retrieval accuracy (see Sec.~\ref{app:video-text-supervision}), we further investigate whether native visual supervision provides advantages for specific visual semantics.

\paragraph{Retrieval-level analysis.} On MSR-VTT, the text-only and video-supervised models achieve 54 and 59 exclusive top-1 retrieval successes, respectively. Among queries where either model fails at rank 1, the ground-truth video still appears within ranks 2--5 in 83--89\% of cases, suggesting that most errors reflect relatively small ranking differences. We further manually examine the 113 exclusive successes for action verbs, spatial relations, and temporal expressions. These categories are not underrepresented among the text-only model's successes, with spatial and temporal cues appearing more frequently in this group. Thus, we find no aggregate evidence that text-only supervision is selectively weaker on visually demanding queries.

\paragraph{Similarity-level analysis.} To examine finer-grained representational differences, we conduct a controlled probe using 20 held-out VideoUFO videos with detailed captions containing visual attributes, spatial relations, and temporal dynamics. For each caption, we construct three manually verified semantic edits and measure their effect on video--text cosine similarity. Altering fine-grained visual attributes produces comparable similarity reductions for the text-only and video-supervised models (0.07 vs. 0.08). In contrast, modifying event order yields a smaller reduction for the text-only model (0.02 vs. 0.04), suggesting a modest advantage of native video supervision in capturing temporal ordering. Spatial-relation edits have limited effects on both models.

Overall, these findings suggest that text-only supervision effectively preserves much of the visual information encoded in the pretrained MLLM, while native video supervision may better capture certain temporal distinctions. Such differences can remain obscured by standard retrieval metrics, as captions often contain multiple redundant identifying cues.

\subsection{Reranking}
Our full model includes an inference-time reranking stage, bares additional computational cost through additional forward passes over the top-K candidates. This may limit applicability for large K values. Furthermore, our reranking relies on a simple pairwise scoring using the MLLM head, and exploring more advanced reranking strategies remains future work.

\section{Societal Impact}
In our method, biases may be inherited both from the pretrained MLLM backbone and from the text-only alignment data. Since our in-context optimization relies on dense video captions and short summaries without visual supervision, hallucinations, omissions, or stereotypes in the captions may propagate into the embedding space and affect retrieval fairness.

\section{Licenses}
We leverage existing resources, namely the VideoLLaMA3 model and the VideoUFO dataset. VideoLLaMA3 is released under the Apache 2.0 license, while VideoUFO is released under the CC BY 4.0 license. We use these resources in accordance with their respective license terms.